\documentclass[11pt,twocolumn,letterpaper]{article}

\usepackage{wacv}
\usepackage{times}
\usepackage{epsfig}
\usepackage{graphicx}
\usepackage{amsmath}
\usepackage{amssymb}

\usepackage{amsmath,amssymb,amsfonts}
\usepackage{algorithmic}
\usepackage{caption}
\usepackage{graphicx}
\usepackage{multirow}
\usepackage{textcomp}
\usepackage{xcolor}
\usepackage{subcaption}
\usepackage{enumitem}

\allowdisplaybreaks

\newcommand{\B}[1]{\boldsymbol{#1}}

\newcommand{\norm}[1]{\left\lVert#1\right\rVert}
\newlength{\tempheight}
\newlength{\tempwidth}

\newcommand{\columnname}[1]
{\makebox[\tempwidth][c]{\textbf{#1}}}

\newcommand{\rowname}[1]
{\rotatebox{90}{\makebox[\tempheight][c]{\bf #1}}}

\def\wacvPaperID{805} 

\wacvfinalcopy 
\ifwacvfinal
\def\assignedStartPage{9876} 
\fi

\ifwacvfinal
\usepackage[breaklinks=true,bookmarks=false]{hyperref}
\else
\usepackage[pagebackref=true,breaklinks=true,colorlinks,bookmarks=false]{hyperref}
\fi

\ifwacvfinal
\else
\fi

\begin{document}

\title{Adversarial Robustness of Deep Sensor Fusion Models}

\author{Shaojie Wang, Tong Wu, Ayan Chakrabarti, Yevgeniy Vorobeychik\\
Washington University in St. Louis\\
{\tt\small \{joss, tongwu\}@wustl.edu, ayan.chakrabarti@gmail.com, yvorobeychik@wustl.edu}

}

\maketitle

\begin{abstract}
We experimentally study the robustness of 
deep camera-LiDAR fusion architectures for 2D object detection in autonomous driving.
First, we find that the fusion model is usually both more accurate, and more robust against single-source attacks than single-sensor deep neural networks.
Furthermore, we show that without adversarial training, early fusion is more robust than late fusion, whereas the two perform similarly after adversarial training.
However, we note that single-channel adversarial training of deep fusion is often detrimental even to robustness.
Moreover, we observe cross-channel externalities, where single-channel adversarial training reduces robustness to attacks on the other channel.
Additionally, we observe that the choice of adversarial model in adversarial training is critical: using attacks restricted to cars' bounding boxes is more effective in adversarial training and exhibits less significant cross-channel externalities.
Finally, we find that joint-channel adversarial training helps mitigate many of the issues above, but  does not significantly boost adversarial robustness.
\end{abstract}

\section{Introduction}
\label{sec:intro}

\begin{figure}[h]
\setlength{\tempwidth}{.3\linewidth}
\settoheight{\tempheight}{\includegraphics[width=\tempwidth]{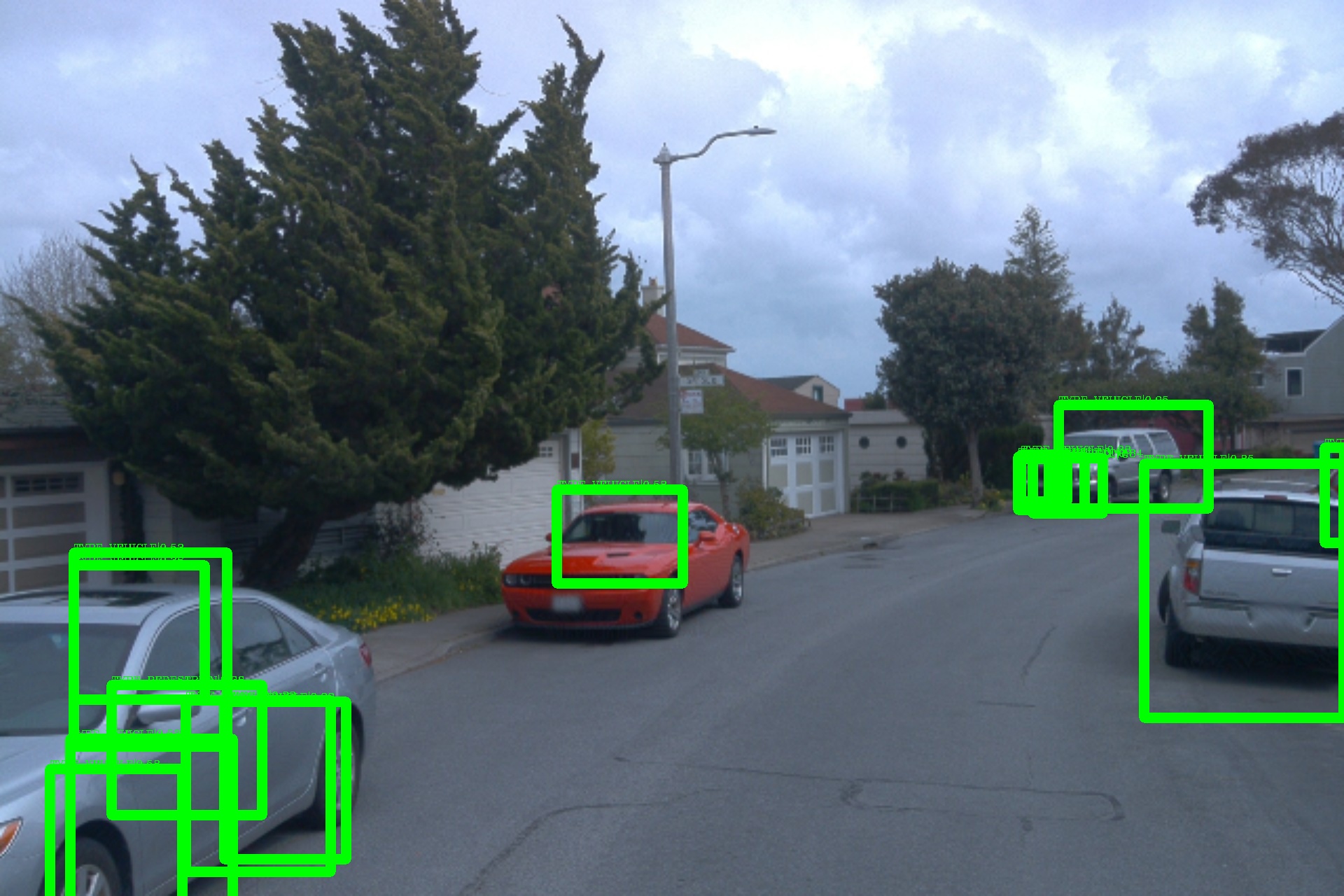}}%
\centering
\hspace{\baselineskip}
\columnname{Clean data}\hfil
\columnname{Full-image}\hfil
\columnname{Car boxes}\\
\rowname{Original}
\subfloat{\includegraphics[width=\tempwidth]{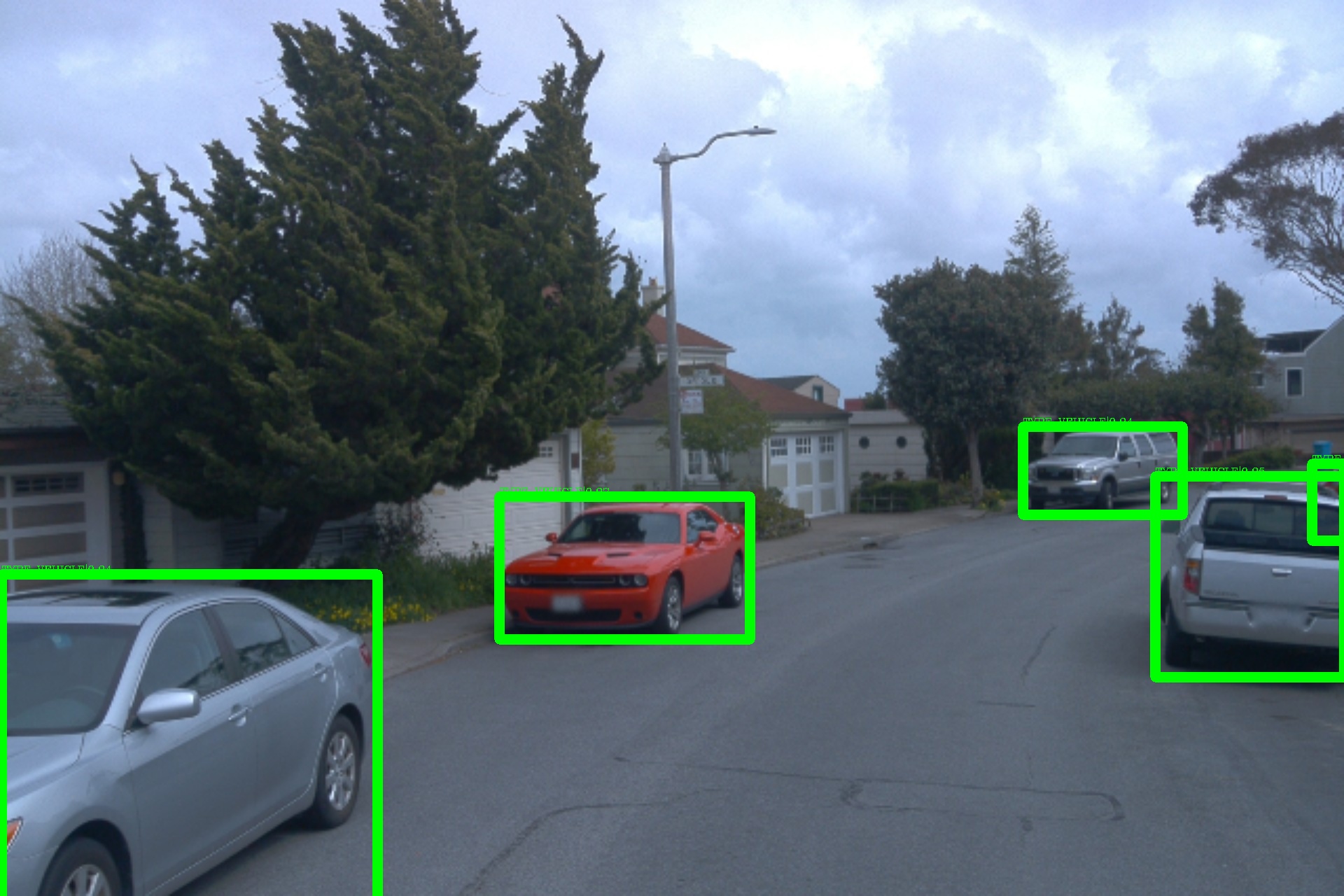}}\hfil
\subfloat{\includegraphics[width=\tempwidth]{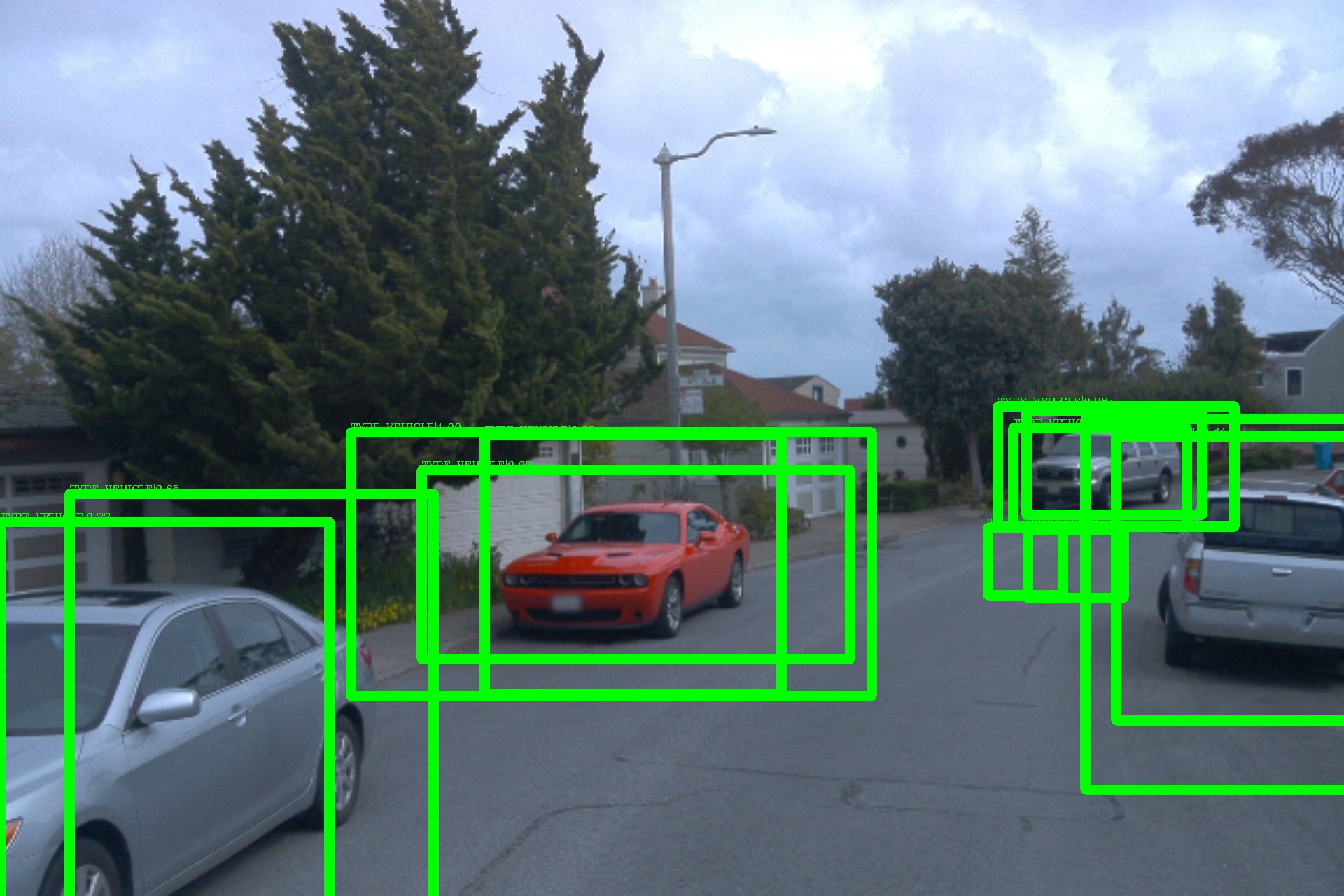}}\hfil
\subfloat{\includegraphics[width=\tempwidth]{LaTeX/Figure/late_org_advcar.jpg}}\\
\rowname{AT-Image}
\subfloat{\includegraphics[width=\tempwidth]{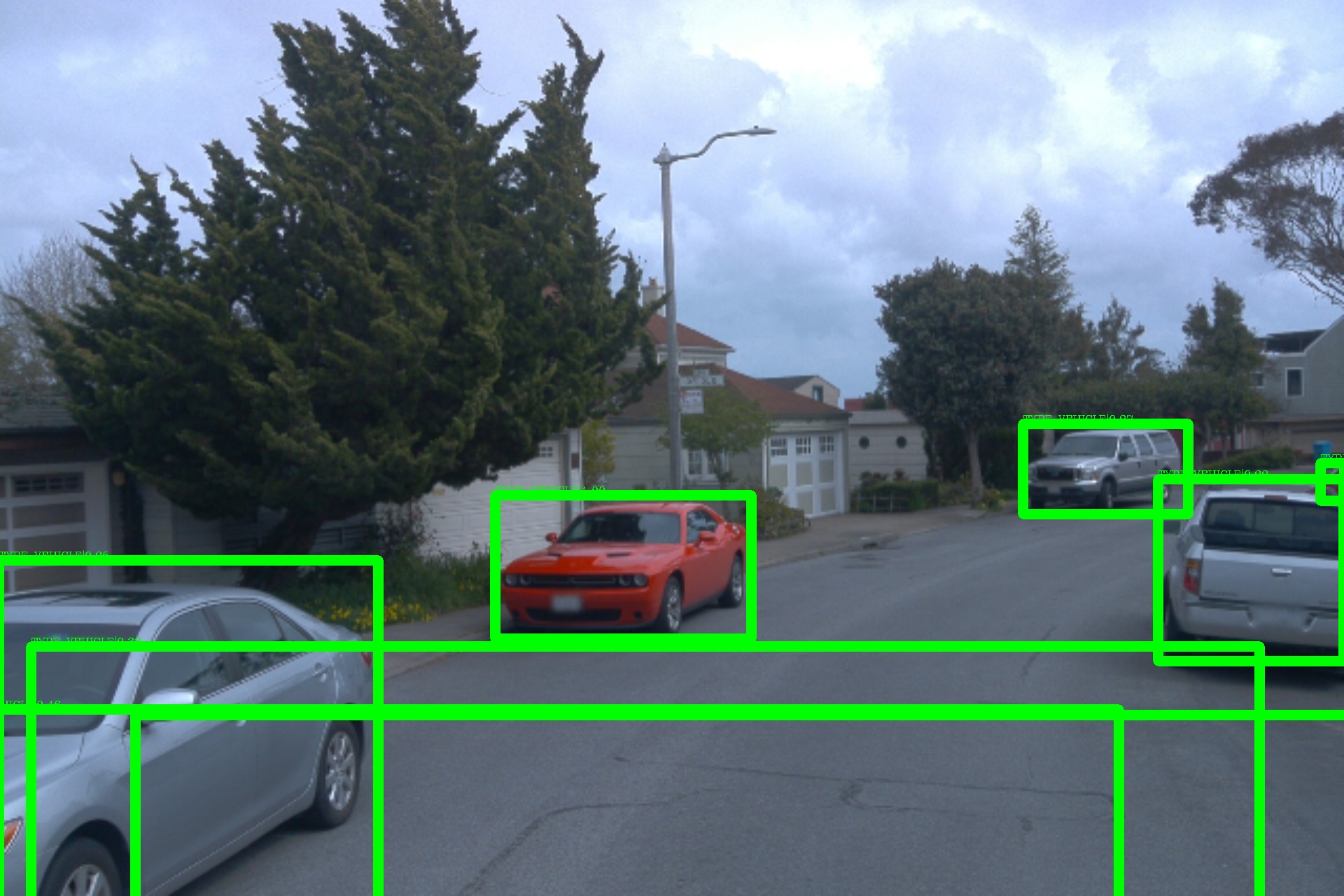}}\hfil
\subfloat{\includegraphics[width=\tempwidth]{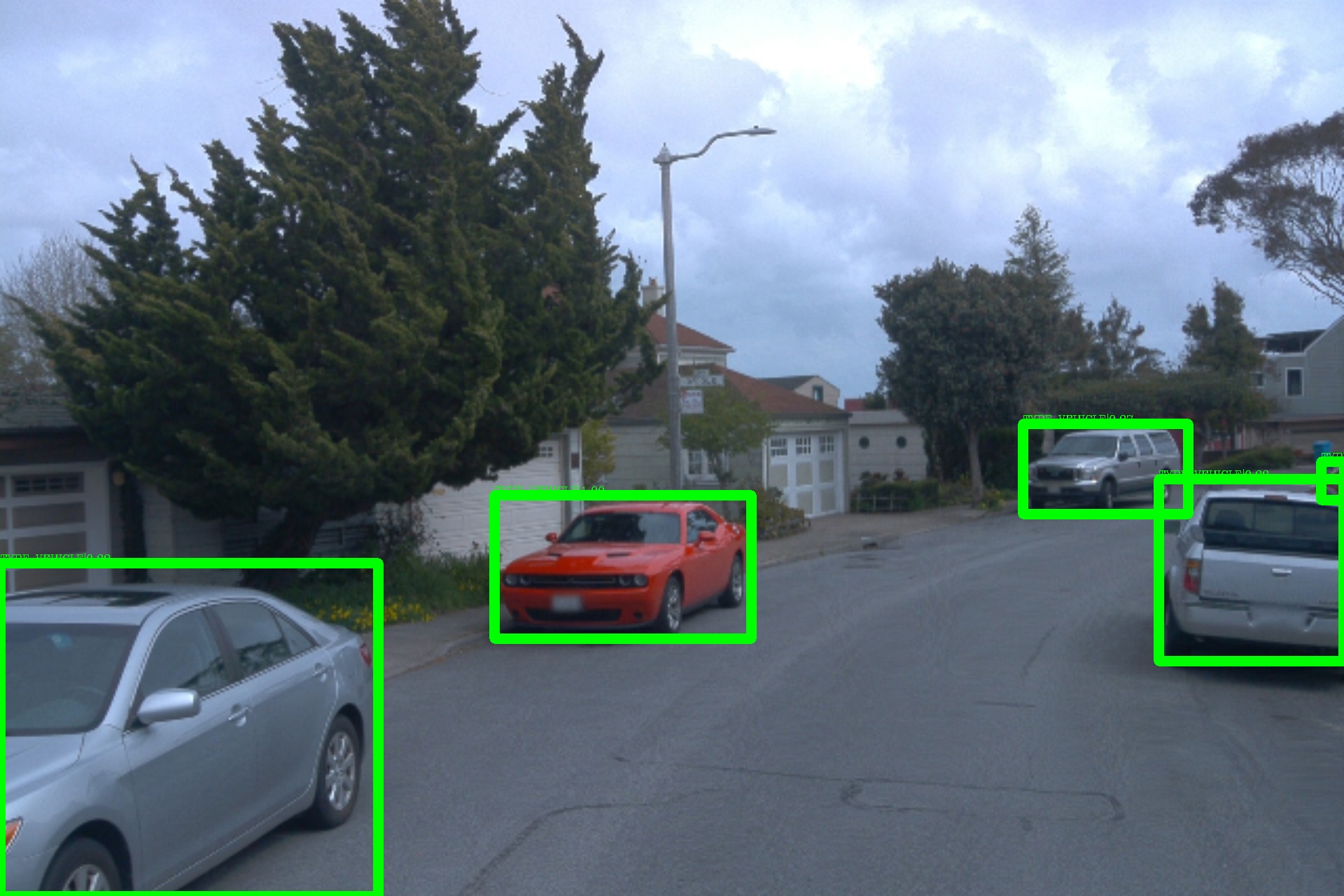}}\hfil
\subfloat{\includegraphics[width=\tempwidth]{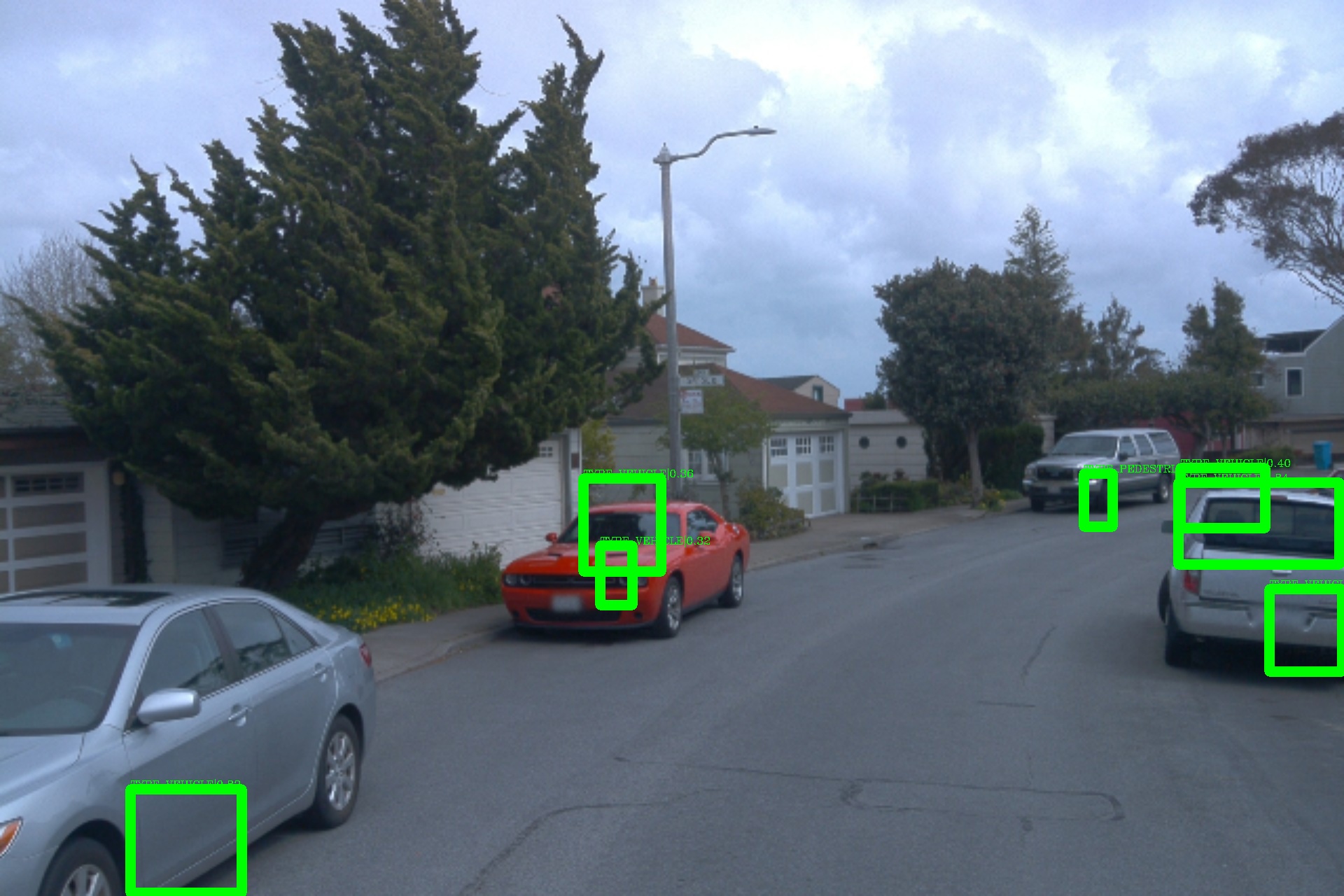}}\\
\rowname{AT-Car\textcolor{white}{g}} 
\subfloat{\includegraphics[width=\tempwidth]{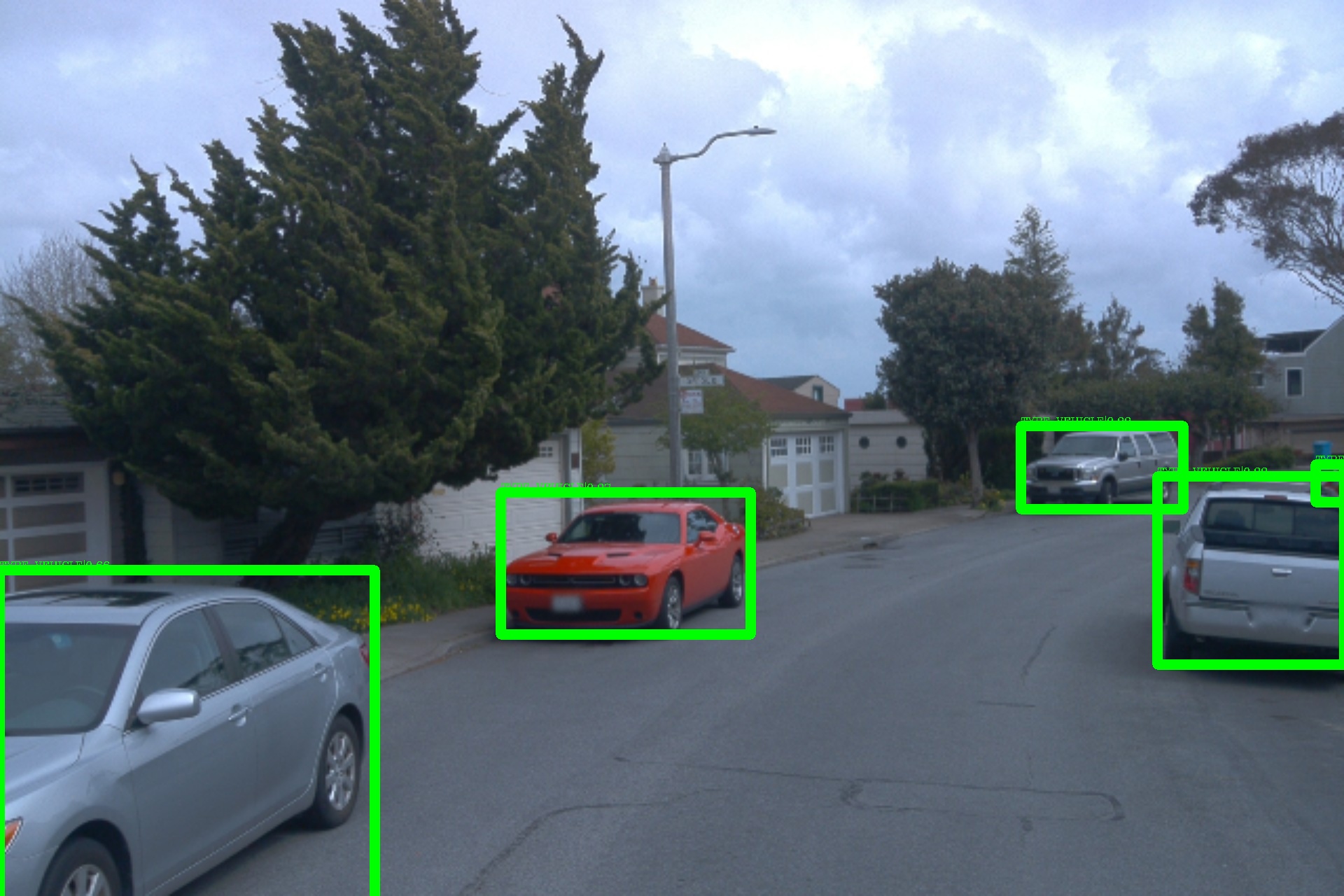}}\hfil
\subfloat{\includegraphics[width=\tempwidth]{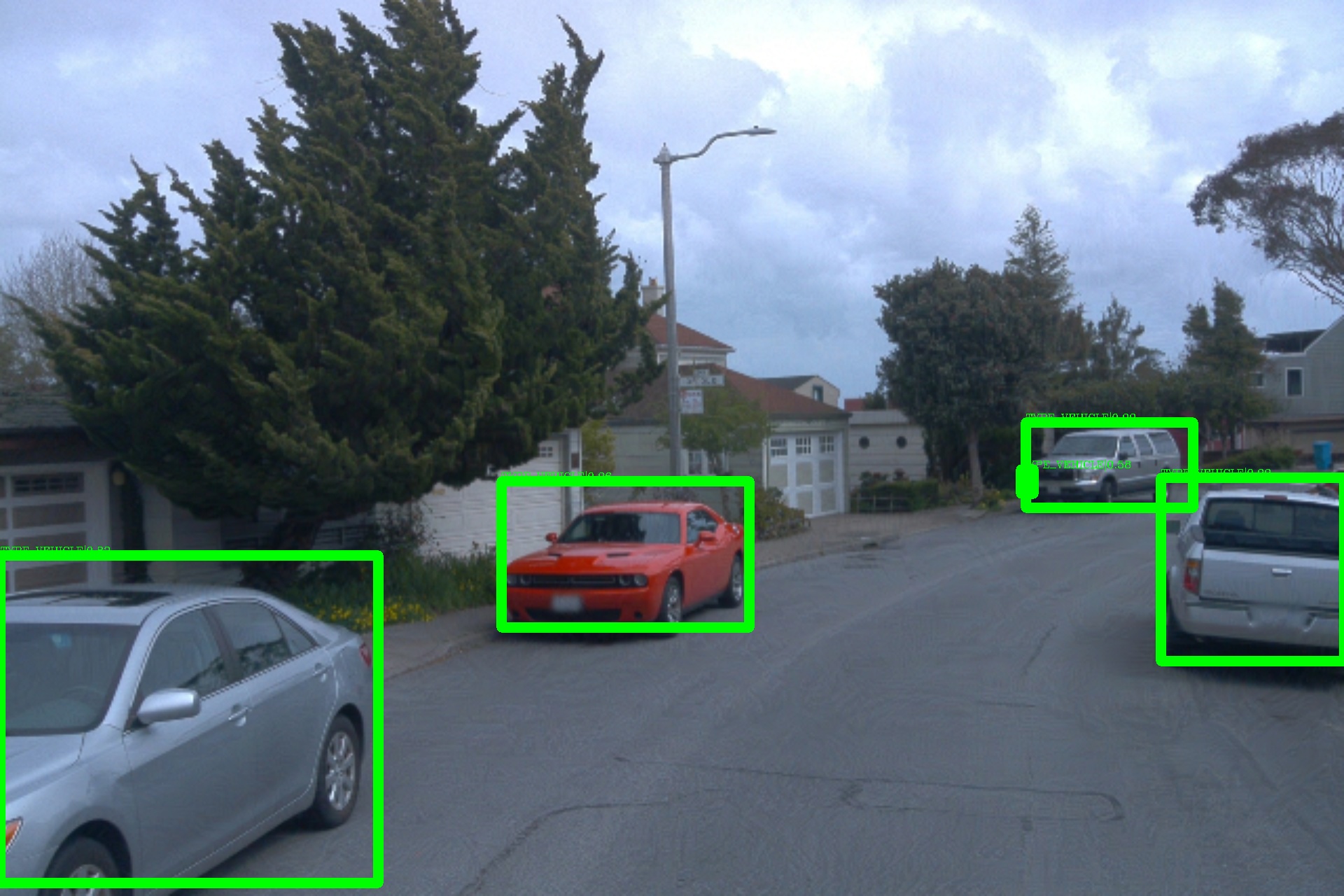}}\hfil
\subfloat{\includegraphics[width=\tempwidth]{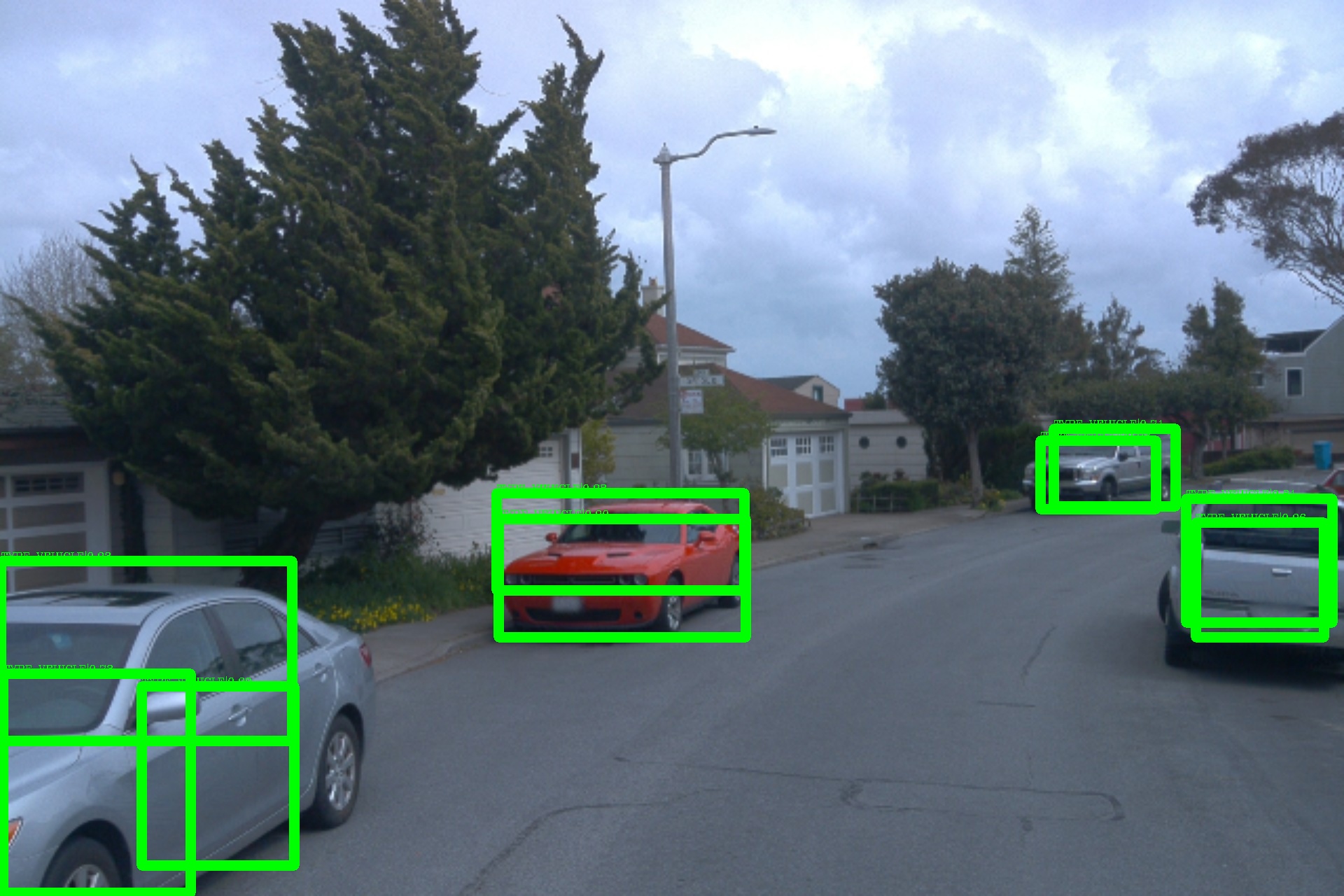}}
\caption{Visualization of late fusion 2D detection results.  Top row: original late fusion model.  Middle row: late fusion after adversarial training with full-image attacks.  Bottom row: late fusion after adversarial training with attacks restricted to cars' bounding boxes.  Left column: clean data.  Middle column: full-image attacks.  Right column: car bounding box attacks.  Note the image in middle row, left column, which demonstrates "adversarial overfitting", that is, far more conservative bounding box generation after full-image adversarial training.}
\label{fig:vis-overfit}
\end{figure}

Autonomous driving (AD) depends heavily on visual perception, such as camera, radar, and LiDAR.
While deep neural network architectures have been transformative in improving automated perceptual efficacy in tasks such as classification and object detection, a series of demonstrations have shown that these perceptual frameworks are also vulnerable to small adversarial perturbations.
While much of the attention has been devoted specifically to attacks on image-channel perception~\cite{advimg-sign3,advimg-sign1,advimg-sign2,fgsm,cvprw20,confadv}, there is now a great deal of evidence that the LiDAR channel is also vulnerable~\cite{advpts-perception,Sun20,Tu20}.

The fusion of information from the myriad of available sensors, however, appears to be an opportunity for creating both more effective and more robust perceptual systems by integrating complementary information.
For example, a LiDAR point cloud can provide additional depth information for RGB images.
However, sensor fusion models are also not without limitations.
For example, suboptimal fusion can lead to performance degradation even compared to single-channel models~\cite{rsf-avsr}.
Additionally, several efforts have shown that  fusion models can exhibit single-channel and multi-channel vulnerabilities~\cite{advfusion,cvprw20,rgbd2-avod}.

However, research so far has largely focused on identifying vulnerabilities in specific fusion architectures.
We take a broader perspective and consider robustness of sensor fusion along four dimensions:
1) comparing robustness of deep sensor fusion to  single-input-channel neural networks in the context of single-channel attacks,
2) the impact of fusion architecture on robustness,
3) the impact of the nature of the threat model, and
4) the efficacy of adversarial training (AT) in improving robustness to single-channel attacks.
Regarding AT specifically, recent work by Kim and Ghosh \cite{single-source} showed that single-channel adversarial training may be ineffective, whereas adversarial training with joint-channel attacks yields robustness to \emph{single-channel} attacks.
However, their evaluation only considered Gaussian noise with 
and used the AVOD architecture.
Our investigation, in contrast, is focused on adversarial examples, and is far broader in scope.


We consider deep sensor fusion models with two perceptual input channels: camera (RGB image) and LiDAR (point cloud).
Our analysis is in the context of 2D object detection.
Our focus on 2D rather than 3D detection enables a direct comparison in robustness between single-channel models such as camera-only deep neural networks and sensor fusion models (dimension (1) above).
Furthermore, we construct fusion architectures using YOLOv4 2D detection architecture as the anchor.
Specifically, we compare two general fusion paradigms: early fusion (YOLO-Early), in which the two inputs are fused immediately at the input layer of the model architecture, and late fusion (YOLO-Late), in which we fuse (concatenate) the \emph{features} extracted from each channel input (followed by additional neural network layers), where feature extraction paths are independent in the neural network architectures (dimension (2) above).
For either early or late fusion, as well as a LiDAR-based single-channel model, we use a \emph{depth} map extracted from the point cloud as the neural network input.
This, again, enables direct comparison across architectures. 
We also study AVOD model, which uses a two-stage detection framework, in contrast of single-stage YOLO architectures.

We consider three threat models for each channel: the first one mirrors typical digital attacks where the entire input can be perturbed (for example, the typical $l_\infty$-bounded adversarial example attacks~\cite{pgd}); the second one is meant to be more alike physically realizable attacks by masking adversarial inputs to be only within car bounding boxes (analogous to adversarial patch attacks~\cite{patch}); and the third one, black-box version of the first threat model, where attackers has no access to parameters of the detection network. 
This allows us to explore dimension (3) above.
Finally, we study two classes of adversarial training: first, single-channel adversarial training, where adversarial examples are only introduced for a single channel in the adversarial training loop, and second, joint-channel training, in which attacks on both channels are used in training.
This addresses dimension (4) above.


Our main findings are:
\begin{enumerate}[topsep=0pt,itemsep=-1ex]
    \item Sensor fusion models are both more effective \emph{and} more robust against single-source attacks than single-channel models.
    \item Prior to adversarial training, early fusion is typically more robust than late fusion.  This advantage largely disappears after adversarial training.
    \item Single-channel adversarial training often \emph{decreases} robustness to single-channel attacks \emph{on the same channel}.
    \item Single-channel adversarial training exhibits negative cross-channel externalities, decreasing robustness to attacks on the other channel.
    \item Joint-channel adversarial training boosts robustness, but only slightly.
    \item Conventional digital attacks used in adversarial training yield models that are  robust neither to these attacks, nor to the more physically realizable attacks restricted to a car's bounding box.  This appears to be a form of adversarial overfitting (the model becomes too conservative), and is illustrated in Figure~\ref{fig:vis-overfit}.  Adversarial training with restricted attacks is more effective and has fewer deleterious side-effects.
\end{enumerate}


\section{Related Work}
\label{sec:related}
\noindent{\bf Robust Sensor Fusion } 
Data collected from multiple sensors are fused together to deal with complex tasks, e.g., audio-visual fusion in speech recognition (e.g.~\cite{avsr1,avsr2}) and video captioning (e.g.~\cite{videocap1,videocap2}). We refer the readers to~\cite{FengHaase2019multimodal} for a detailed literature review on AD sensor fusion.
As for robust sensor fusion, \cite{rsf-avsr} studies the robustness of audio-visual fusion against catastrophic fusion problem in speech recognition. \cite{rsf-multisensordet} proposes a multimodal model which is robust to the absence of some sensors. \cite{rsf-uncertainty} focuses on the uncertainty of input data from different sensors, including the differences in physical units of measurement, sampling resolutions, and spatio-temporal alignment.~\cite{single-source} are the first to mention adversarial robustness of sensor fusion models, but  only provide a theoretical proof for robustness of logistic regression against single-source adversarial noise.~\cite{robust-multi-sensor} 
experimentally study AT to achieve robust sensor fusion, but use highly visible attacks.

\noindent{\bf Adversarial Perturbations }
Adversarial perturbations add small noise to sensor inputs~\cite{pgd,fgsm}.
Recently, physically realizable attacks have been proposed in the context of AD. 
For example, a number of attacks add stickers or adversarial patches to objects, such as stop signs and road pavement, or otherwise perturb these, in order to cause AD mistakes~ \cite{advimg-sign1,advimg-sign2,advimg-sign3,advimg-sign4,advimg-steerangle,patch,advimg-person}.
Similarly, attacks on point cloud input modalities range from perturbing all points~\cite{advpts-points} to physically realizable attacks that spoof the LiDAR sensor~\cite{advpts-perception} and introducing real objects that cannot be detected by a LiDAR~\cite{advpts-object}.
However, relatively few approaches study attacks specifically against fusion architectures.
The few that do only consider the image inputs~\cite{advfusion,cvprw20}, or consider a single target model, such as AVOD~\cite{robust-multi-sensor}.
\section{Attacking and Defending Sensor Fusion for Object Detection}
\label{sec:models}

We begin by describing the object detection framework, and the use and several forms of sensor fusion in this context.
We then present a generalization of single-sensor attacks to apply to both multiple input modalities in fused models, as well as to model digital and physically realizable attacks.
Finally, we describe natural variations of adversarial training as a means to defend object detection models that utilize sensor fusion from single-channel and multi-channel attacks.

\subsection{Sensor Fusion for Object Detection}

In addition to studying the AVOD~\cite{rgbd2-avod} sensor fusion architecture specifically, 
we are investigate the effect of different fusion paradigms on adversarial robustness. For this, we select YOLOv4~\cite{yolov4}, a single-stage detector, as our base model, and design two variations of YOLOv4 - YOLO-Early and YOLO-Late, where YOLO-Early fuses the two sensor inputs immediately at the beginning of the model, and YOLO-Late delayes the fusion operation until the feature output layer.

For consistency, we convert the input point cloud representation into a depth map, which is then used as the LiDAR sensor input into deep neural networks.
Additionally, we project each point of the depth map onto the RGB image, and use nearest interpolation to generate a \emph{dense depth map}, which serves as one LiDAR input channel.
In addition, we have a second LiDAR channel, the \emph{distance map}, which records the distance between the interpolated point and the original point. 
Since the two LiDAR channels are pixelwise consistent with the RGB images, we can either add them as a fourth and fifth channel, resulting in our RGBD representation, or use the two LiDAR channels as independent LiDAR inputs.
We call the RGBD input representation \emph{early fusion} (and the corresponding YOLO variant \emph{YOLO-Early}), since both image and LiDAR inputs are combined early in the deep neural network architecture.
Our \emph{late fusion} variants (referred to as \emph{YOLO-Late}) use separate feature extraction routes for RGB and two-channel LiDAR-based depth and distance maps, with the extracted features being fused later in the architectural structure of the neural network. 
Finally, we can also use single-sensor variants, such as RGB-input-only and LiDAR (depth and distance)-input-only models (referred to as \emph{YOLO-RGB} and \emph{YOLO-Depth} respectively).
We implement these architectures using YOLOv4.
Specifically, \emph{YOLO-Early} directly feeds the 5-channel RGBD image into the original YOLOv4 model. \emph{YOLO-Late} uses two separate feature extraction routes for the two channels, each has its individual Darknet backbone and feature pyramid network (FPN) neck. Then, we concatenate both features and subsequently feed them into the shared bounding box head for final prediction.

All models are trained to detect three classes of objects: cars, pedestrians, and cyclists.
We use mean average precision (mAP) to evaluate the effectiveness of object detection, both with and without attacks, where significant reduction in mAP after an attack entails a successful attack.

\subsection{Attacks on Sensor Fusion for Object Detection}
\label{S:attacks}

We consider white-box decision-time (adversarial perturbation) attacks on inputs into the neural networks that combine data from a collection of $n$ sensors.
To formalize, let $f(\boldsymbol{x}_1, \boldsymbol{x}_2, \dots, \boldsymbol{x}_n; \B{\theta})$ be a neural network model for object detection that takes inputs $\boldsymbol{x}_i$ from the $n$ sensors.
While our specific focus below will be on only two sensor modalities, RGB image and LiDAR (represented by a distance and depth maps, as discussed above), we describe the adversarial framework more generally.
The learner's goal is to learn model parameters $\B{\theta}$ that minimize the prediction loss $L(f(\boldsymbol{x}_1, \boldsymbol{x}_2, \dots, \boldsymbol{x}_n; \B{\theta}),y)$, and we model the adversary's goal as maximizing this loss \emph{for a fixed vector of model parameters} $\theta$.
Let $\mathcal{I}$ denote the the set of indeces of sensors which are attacked, and $\mathcal{I}^c$ the indeces of sensors which are not (with $\mathcal{I} \cup \mathcal{I}^c = [1..n]$, where $[1..n]$ is the set of integers between 1 and $n$, and $\mathcal{I} \cap \mathcal{I}^c = \emptyset$).
A natural generalization of adversarial example attacks to the sensor fusion setting is then


\begin{align}
\label{eq:objective}
    \underset{\boldsymbol{\delta}}{\arg \max }& \  L(f(\dots, \boldsymbol{x}_i + \boldsymbol{\delta}_i * \boldsymbol{m}_i, \dots ; \boldsymbol{\theta}), y) \quad \\
    s.t.\ \ \  
         &  \norm{\boldsymbol{\delta}_i}_p \leq \epsilon_i, \forall i \in \mathcal{I}, \ \boldsymbol{\delta}_i = 0 , \forall i \in \mathcal{I}^c, \nonumber
\end{align}
where 
$\boldsymbol{\delta}_i$ is the perturbation on the $i$-th sensor input, $\epsilon_i$ is the bound on the $l_p$ norm perturbation of the $i$th sensor attack, 
and $\boldsymbol{m}_i$ is a mask which constrains the attack on sensor $i$ to an exogenously specified contiguous region of the input (e.g., image or LiDAR).
In our setting, perturbations to images take the form of modifying pixel values for the three color channels.
For LiDAR, in contrast, we modify the input \emph{point cloud} (depth and distance maps are thereby indirectly affected) by shifting each point in 3D space.

The idea behind introducing masks $\B{m}_i$ is that it allows us to capture a common distinction between \emph{digital} and \emph{physically realizable} attacks~\cite{Wu20}.
In particular, we will consider digital attacks, in which the mask is simply the entire input (i.e., no mask), as well as physically realizable attacks in which the mask is restricted to the bounding boxes of cars in scene.
As our focus is on image and LiDAR input modalities, and since perturbations to these inputs are not directly comparable, we emphasize this by using $\epsilon$ to refer to the bound on the RGB inputs (denominated in pixel color value intensities) and $\gamma$ to denote the bound on LiDAR point cloud perturbations (denominated in meters).
Throughout, we focus discussion on $l_\infty$-norm perturbations.
We use the $l_\infty$ variant of projected gradient descent (PGD) for all attacks, on both the image and LiDAR sensor modalities.



In object detection one can have several possible objectives in adversarial perturbations, such as confidence scores~\cite{confadv} and regression loss~\cite{bestadv}.
We focus on the latter, which is highly effective.

\subsection{Defense through Adversarial Training}
\label{S:AT}

A common and usually most effective way to obtain increased robustness to adversarial perturbation attacks is \emph{adversarial training}.
A common mathematical framework for adversarial training is robust optimization of the following form:
\begin{align}
     \boldsymbol{\theta}^* = \underset{\boldsymbol{\theta}}{\arg \min }\ \mathbb{E}[ \max_{\boldsymbol{\delta}} L(f(\dots, \boldsymbol{x}_i + \boldsymbol{\delta}_i * \boldsymbol{m}_i, \dots ; \boldsymbol{\theta}), y)]
\end{align}
Learning is then a form of gradient descent with respect to $\B{\theta}$, evaluated at the value of $\B{\delta}$ which approximately maximizes the inner optimization of Problem~\eqref{eq:objective}.
While PGD has commonly been used to obtain the approximate solution to the inner adversarial optimization problem~\cite{pgd}, we utilize a recently proposed fast alternative that uses FGSM with random initialization and cyclic learning rate~\cite{fast}.

We utilize two forms of adversarial training.
The first form only adds adversarial perturbations for a single input channel, with the aim of making that channel only more robust to attack.
The second form adds perturbations to both input channels (image and LiDAR) concurrently in the adversarial training loop.
Below we study the relative efficacy of these two variations.
In addition, we consider two variations of single-channel threat models in adversarial training: digital attacks (i.e., adversarial training using attacks without a mask) and physically realizable attacks (in which adversarial perturbations are restricted to a small contiguous region of the input).

\section{Experiments}
\label{sec:exp}


\subsection{Experiment Setup}

\noindent{\bf Data }
We perform experiments on Waymo~\cite{waymo} and KITTI~\cite{kitti} datasets. Since the results on both datasets are highly consistent, we only present the results on Waymo in the main text, with KITTI results deferred to the supplement.
Waymo has 5 cameras in total. In this work, we only use the front camera for images. We select front images and LiDAR point clouds evenly from $1/10$ of all the frames for training. For testing, we select the first frame from each sequence in the validation set, which gives us 202 samples in total. The resemblance among different frames within the same sequence helps reduce the effect of sampling on the evaluation performance to the minimum.

\noindent{\bf Training }
When training YOLOv4 models, we start everything from scratch, i.e. no pretrained weights are used. This is to avoid the problem where models pretrained on other image datasets might make the model biased to images and ignore the LiDAR sensor in 2D detection. More details can be found in the supplementary. For AVOD model, we follow the instructions in the original paper to train it on KITTI.

%
\noindent{\bf Attacks and Defenses} We consider three attacks on both image and LiDAR channels - \emph{digital} attacks that perturb the entire input, \emph{physically realizable} attacks, and \emph{black-box} digital attacks, which are described in Section~\ref{S:attacks}. For defenses, we consider variants of adversarial training (AT), described in Section~\ref{S:AT}, based on each of the four single-sensor attacks: \emph{AT-image}, which uses the full-image $l_\infty$ attack, \emph{AT-LiDAR}, which attacks the full point cloud, as well as \emph{AT-Car} and \emph{AT-LiDAR-Car}, which use attacks within the car bounding boxes only.
Additionally, we consider adversarial training using attacks on \emph{both} sensor modalities jointly, referred to as \emph{AT-Joint}.
For adversarial training on the image channel, $\epsilon = 2$, while adversarial training on the LiDAR channel uses $\gamma = 0.3$.


\subsection{Adversarial Robustness of Fusion Models}

Our first question involves relative performance of LiDAR-image fusion models in comparison with single-sensor models.
First, we compare the effectiveness of early and late fusion models on \emph{clean}, that is, original unperturbed, test data.
It is, of course, not surprising that both early fusion (0.361)\footnote{Numbers in parenthesis following the model name denotes its mean average precision (mAP) performance.} and late fusion (0.350) models significantly outperform both the image-only (0.322) and LiDAR (depth)-only (0.238) single-channel models.

\begin{figure}[h]
    \centering
    \begin{subfigure}{0.235\textwidth}
    \centering
    \captionsetup{justification=centering}
    \includegraphics[width=\textwidth]{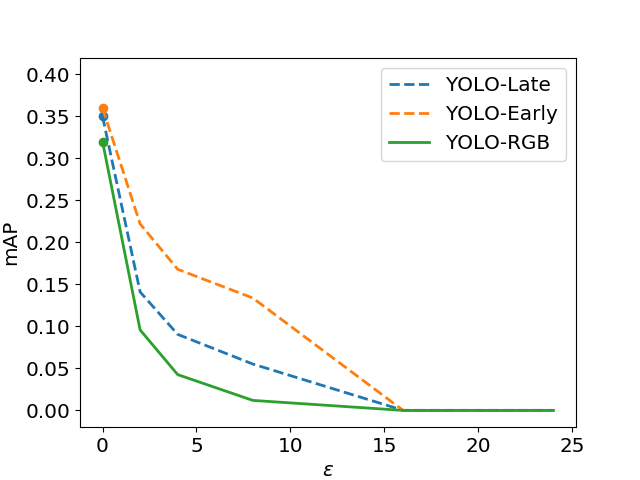}
    \caption{full-image attack}\label{fig:yolo-og-imageatk}
    \end{subfigure}
    \begin{subfigure}{0.235\textwidth}
    \centering
    \captionsetup{justification=centering}
    \includegraphics[width=\textwidth]{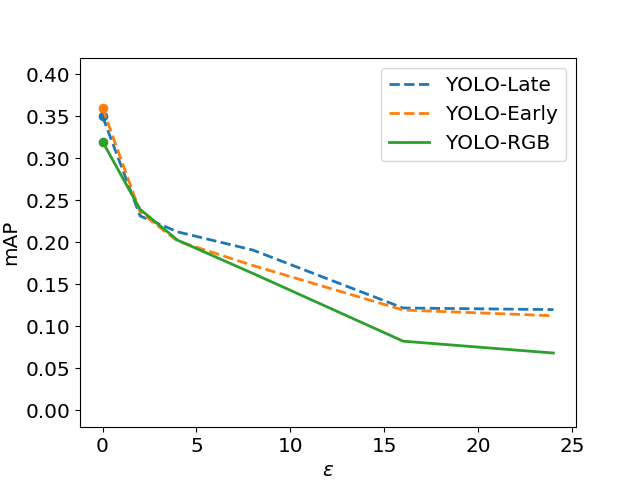}
    \caption{car-image attack}\label{fig:yolo-og-carimg}
    \end{subfigure}
    \begin{subfigure}{0.235\textwidth}
    \centering
    \captionsetup{justification=centering}
    \includegraphics[width=\textwidth]{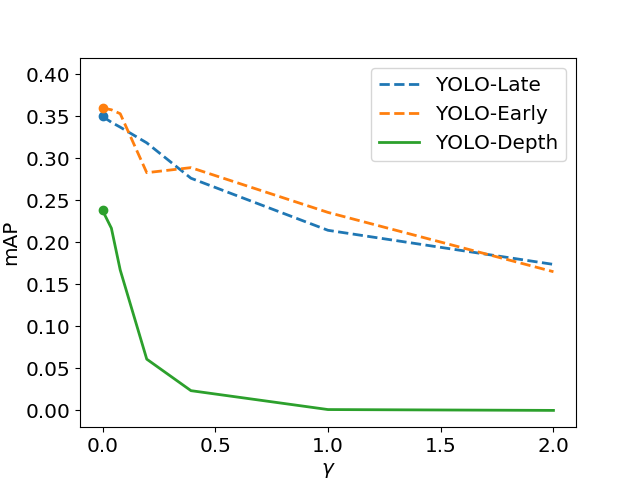}
    \caption{full-LiDAR attack}\label{fig:yolo-og-lidaratk}
    \end{subfigure}    
    \begin{subfigure}{0.235\textwidth}
    \centering
    \captionsetup{justification=centering}
    \includegraphics[width=\textwidth]{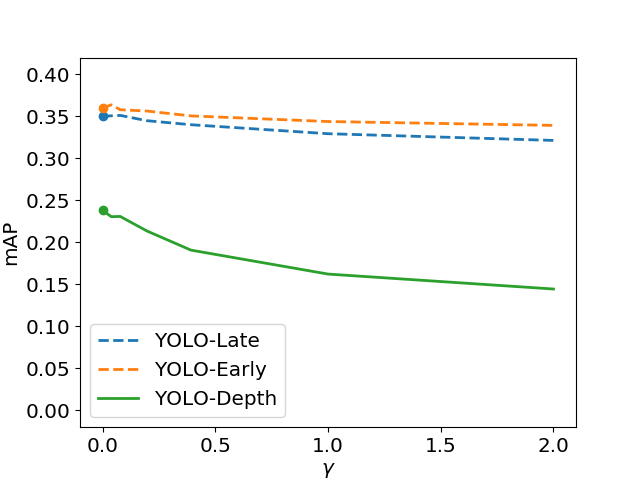}
    \caption{car-LiDAR attack}\label{fig:yolo-og-carlidaratk}
    \end{subfigure}
    \caption{Robustness of deep sensor fusion, compared to single-channel neural network object detection models, to adversarial attacks on image and LiDAR modalities.} 
    \label{fig:imgc}
    
\vspace{-0.4cm}
\end{figure}
Figure~\ref{fig:imgc} offers a more complete comparison in terms of adversarial robustness to our four attacks: the top row presents robustness to image attacks, whereas the bottom row shows robustness to LiDAR attacks.
In all cases, we can see that not merely accuracy is considerably improved by fusion, but also single-channel robustness.
Most significant improvements of fusion models in comparison to single-channel variants are in the context of full-image attacks (where early fusion is much more robust than the image-only model), and in the context of LiDAR attacks, where either fusion variant is far more effective both in clean data accuracy and robustness than the LiDAR-only counterpart.
We can also observe from Figure~\ref{fig:imgc} that early fusion is typically better than late fusion.
In the case of full-image attacks, this advantage is substantial.
However, in the case of the other three attack variants, the difference between early and late fusion is nearly negligible.

\subsection{Effects of Adversarial Training}

\noindent{\bf Adversarial Training and Image Channel Attacks }
Above, we observed that early fusion appears to be somewhat more robust to \emph{full-image attacks} than late fusion, and slightly better even on clean data.
At the same time, we saw that the difference between these approaches is minimal in the case of other attacks we consider.
Furthermore, in fusion models are also more robust than single-sensor counterparts.
We now investigate the impact of adversarial training on these observations.

\begin{figure}[h]
\centering
    \begin{subfigure}{0.235\textwidth}
    \centering
    \captionsetup{justification=centering}
    \includegraphics[width=\textwidth]{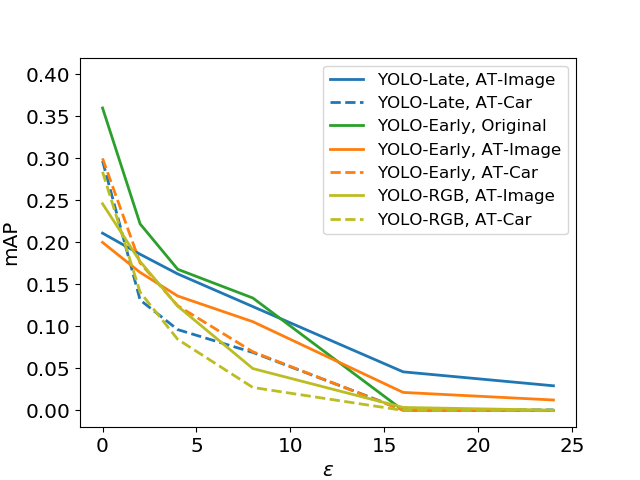}
    \end{subfigure}
    \begin{subfigure}{0.235\textwidth}
    \centering
    \captionsetup{justification=centering}
    \includegraphics[width=\textwidth]{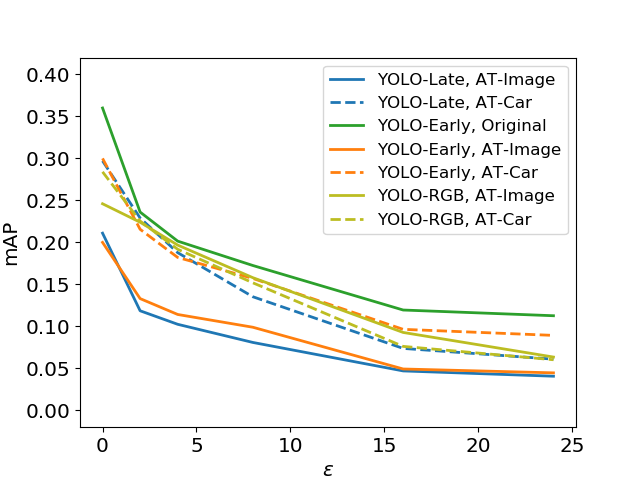}
    \end{subfigure}
    \caption{Early and late fusion after adversarial training, under attacks on  the image channel.  Left: full-image attack.  Right: car bounding box attack.}
    \label{fig:AT-image}
\vspace{-0.4cm}
\end{figure}

In Figure~\ref{fig:AT-image}, we consider both attacks (car bounding boxes only, and full-image), and adversarial training, with respect to the \emph{image} channel.
First, we can note that after adversarial training, the difference between early and late fusion (blue vs.~orange lines) becomes small, whichever adversarial training method is used.
However, if we compare AT-image early fusion between (solid orange lines) and original early fusion (solid green lines), we can see something rather striking: adversarial training actually makes it \emph{less robust}, except for the exceptionally strong attacks (full-image, large $\epsilon$), in which case all models are very bad in any case.
In contrast, adversarial training does boost the robustness of late fusion, although only on the full-image attacks.
Indeed, even if we consider AT-Car models, that is, adversarial training using the weaker attacks inside cars' bounding boxes (dashed lines in Figure~\ref{fig:AT-image}), there seems little improvement in robustness to either type of attack.

A related observation pertains to a comparison between AT-Image, that is, adversarial training using attacks on the full image, and AT-Car, where only attacks within the car bounding boxes are used in training.
We can note that AT-Image (solid lines in Figure~\ref{fig:AT-image}) models are not much more robust than AT-Car models when facing full-image attacks (left plot), but are \emph{much less robust} when faced with car-image attacks.
AT-Image is also significantly worse in performance on clean data, e.g., mAP of early fusion with AT-Image is only 0.200 compared to its original performance 0.361.
Furthermore, in the case of car-image attacks, any advantage of sensor fusion over the image-only model largely dissipates after adversarial training.

Interestingly, the observations above appear to be specific to sensor fusion models.
In particular, if we consider image-only models (YOLO-RGB in the plots), adversarial training has both a comparable or lesser impact on clean accuracy, but also less value for robustness to full-image attacks with a large $\epsilon$.
On the other hand, these image-only models remain comparable to AT-Car variants of sensor fusion, even when trained using full-image attacks.

We can conclude that \emph{the choice of threat model in adversarial training is critical, and conventional digital attacks in which every pixel in an image is suspect are perhaps a particularly poor choice}.
More generally, if the adversarial model is too strong, and likely not a realistic representation of the attacker's capabilities, as is the case with full-image attacks, using it for adversarial training leads one to become too conservative (we can think of this as a form of overfitting to the adversarial model).
This is visualized in Figure~\ref{fig:vis-overfit}, where AT-Image late fusion model yields a very large bounding box for a car: indeed, it is safe in the sense that the bounding box includes the object, but liveness (ability to precisely identify boundaries of an obstacle) is critically compromised.

\noindent{\bf Adversarial Training and LiDAR Channel Attacks }
Next, we explore the effect of full-LiDAR attacks and LiDAR attacks restricted to cars' bounding boxes used both for adversarial training and evaluation (in other words, the threat model is a LiDAR-channel attack, consistent for both attack and defense).

\begin{figure}[h]
\vspace{-0.4cm}
\centering
    \begin{subfigure}{0.235\textwidth}
    \centering
    \captionsetup{justification=centering}
    \includegraphics[width=\textwidth]{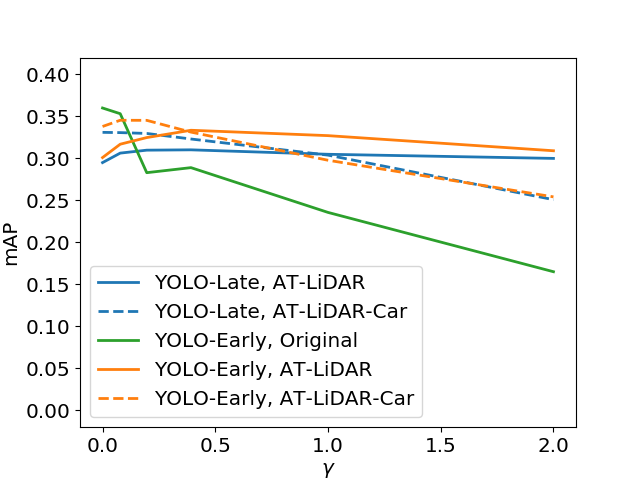}
    \end{subfigure}
    \begin{subfigure}{0.235\textwidth}
    \centering
    \captionsetup{justification=centering}
    \includegraphics[width=\textwidth]{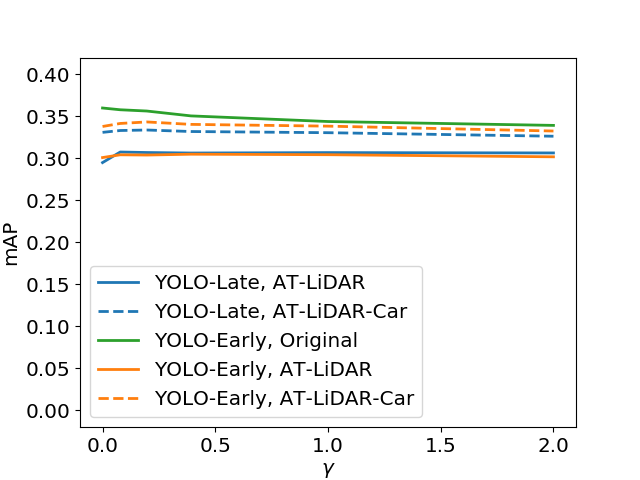}
    \end{subfigure}
    \caption{AT with LiDAR attacks, and LiDAR Channel Attacks. Left: full-LiDAR attack. Right: car-LiDAR attack.}
    \label{fig:AT-lidar}
\end{figure}

Figure~\ref{fig:AT-lidar} presents the results of LiDAR attacks on AT-LiDAR variants. AT for YOLO-Depth is not presented because of its inferior performance.
One interesting difference from our results where attacks are on the image channel is that here we do see a clear improvement in robustness compared to the original early fusion model in the case of full-LiDAR attacks (left plot).
This is true whether adversarial training used the car-LiDAR or full-LiDAR attack model.
Moreover, the difference between AT-LiDAR and AT-LiDAR-Car appears less important than the analogous difference between AT-Image and AT-Car earlier, although AT using LiDAR-Car attacks does yield somewhat better results when evaluated against the same attack. We similarly observe that AT-LiDAR-Car (early fusion: 0.338; late fusion: 0.331) has superior clean data mAP to AT-LiDAR (early fusion: 0.301; late fusion: 0.295).
Nevertheless, we do broadly still observe that the choice of the threat model is important.
Specifically, if we view attacks that are restricted to the cars' bounding boxes (which are more likely to be physically realizable), even the baseline (pre-AT) model is quite robust---in fact, more robust than any of the adversarially trained variants.
Moreover, in this case, adversarial training with full-LiDAR attacks decreases performance the most.

Overall, so far we observe that \emph{single-channel adversarial training appears to be surprisingly ineffective in sensor fusion settings in the context of same-channel attacks}: not only does accuracy on clean data decrease, but even robustness to attack often does.
This appears to be in contrast with single-channel adversarial training, which does increase robustness at least to the threat model used in AT (compare YOLO-RGB between Figures~\ref{fig:imgc}(a) and~\ref{fig:AT-image}(left) for $\epsilon \le 5$).

\noindent{\bf Cross-Channel Externalities of Single-Channel Adversarial Training }
Thus far, we considered only attacks on the same single channel for both adversarial training and robustness evaluation.
We now study a uniquely fusion phenomenon: cross-channel externalities of adversarial training.
Specifically in the context of image and LiDAR channels, we have in mind the following experiments.
On the one hand, we can perform adversarial training with image-channel attacks, but evaluate the resulting model using LiDAR-channel attacks.
On the other hand, we can reverse this, evaluating LiDAR-based AT using image-channel attacks.
The question we ask is: what impact does adversarial training with respect to one channel has on robustness to attacks on the other channel?

\begin{figure}[h]
\centering
    \begin{subfigure}{0.235\textwidth}
    \centering
    \captionsetup{justification=centering}
    \includegraphics[width=\textwidth]{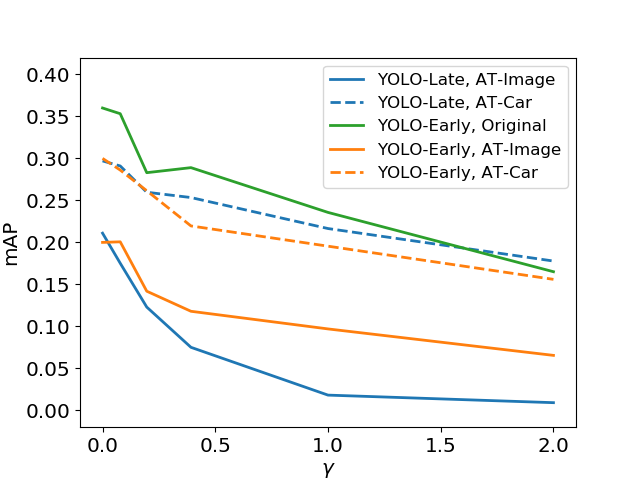}
    \label{fig:at-fulllidar}
    \end{subfigure}
    \begin{subfigure}{0.235\textwidth}
    \centering
    \captionsetup{justification=centering}
    \includegraphics[width=\textwidth]{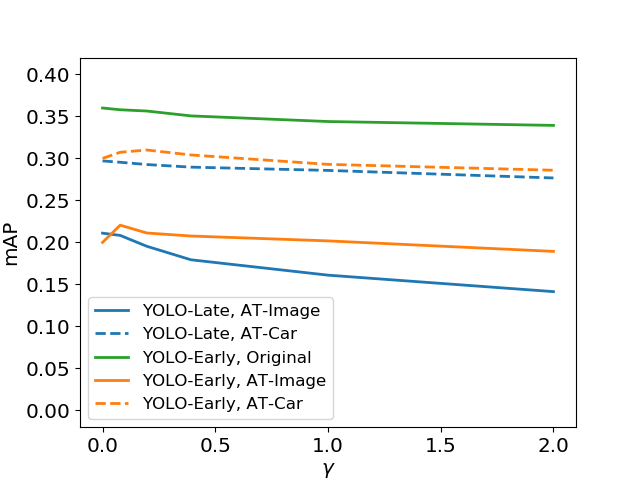}
    \label{fig:at-carlidar}
    \end{subfigure}
    \vspace{-0.6cm}
    \caption{Early and late fusion after adversarial training using image attacks, but subject to LiDAR attacks. Left: full-LiDAR attack. Right: car-LiDAR attack.}
\label{fig:ATI-lidar}
\vspace{-0.4cm}
\end{figure}


\begin{figure}[h]
\centering
    \begin{subfigure}{0.235\textwidth}
    \centering
    \captionsetup{justification=centering}
    \includegraphics[width=\textwidth]{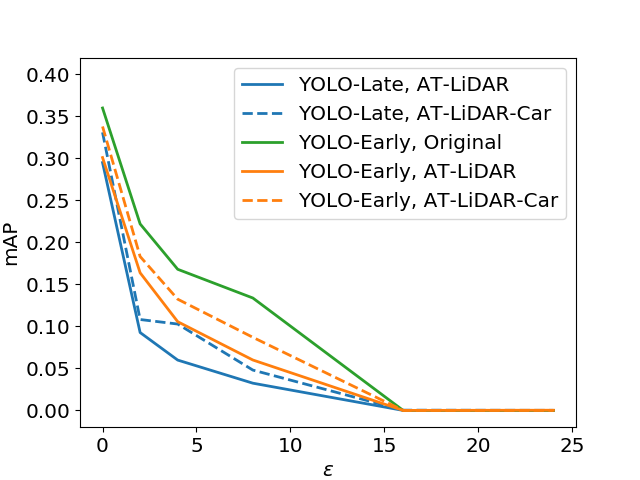}
    \end{subfigure}
    \centering
    \begin{subfigure}{0.235\textwidth}
    \centering
    \captionsetup{justification=centering}
    \includegraphics[width=\textwidth]{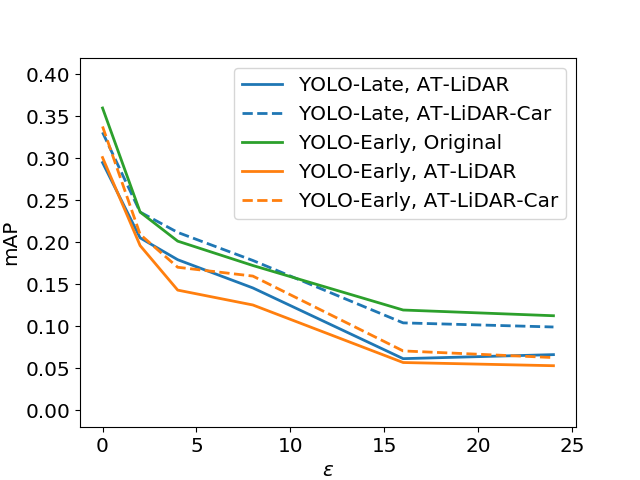}
    \end{subfigure}
    \caption{Early and late fusion after adversarial training using LiDAR attacks, but subject to image attacks. Left: full-image attack. Right: car-image attack.}
\label{fig:ATL-image}
\vspace{-0.4cm}
\end{figure}

The results of this evaluation are in Figure~\ref{fig:ATI-lidar} when we use image attacks for training and LiDAR for evaluation, and in Figure~\ref{fig:ATL-image}, where adversarial training uses LiDAR attacks and evaluation uses image attacks.
First, consider image-based adversarial training in Figure~\ref{fig:ATI-lidar}.
AT with full-image attacks, in particular, causes \emph{dramatic} degradation in both performance on clean data and in robustness to LiDAR-based attacks of either variety.
Interestingly, here we also see that early fusion (solid orange lines) generally does better than late fusion (solid blue lines).
While adversarial training using only the car bounding box adversarial examples exhibits less loss in robustness compared to the model before no adversarial training, there is still a tangible reduction in performance against adversarial examples.
Overall, we see a clear negative externality of image-based adversarial training on robustness to the LiDAR-channel attacks.

Turning to the reverse setting in Figure~\ref{fig:ATL-image}, the results are broadly consistent, but the differences less dramatic than above.
In particular, we still typically see that single-channel adversarial training results in lower performance on the other channel compared to no adversarial training at all.
The one interesting exception is in the performance of AT-LiDAR-Car late fusion model, which is essentially identical (i.e., no cross-channel negative externality) to the original late fusion model (i.e., before AT).

Overall, we therefore find that there are negative cross-channel externalities for both channels.
However, here, too, we see that there is a great deal of importance in the choice of the threat model.
For example, to the extent that we see externalities, they are much stronger when we train with adversarial examples generated on the entire image or point cloud, than AT using only attacks within cars' bounding boxes.

\noindent{\bf Joint-Channel Adversarial Training }
Given the broad failure of AT that we had observed above in inducing much robustness in deep image-LiDAR fusion models, it is natural to consider this failure a product of training with only single-source attacks.
One can therefore hypothesize that all will be well once we adversarially train with attacks on \emph{both} channels simultaneously.
Indeed, precisely this remedy has been proposed in a largely theoretical framework by Kim and Ghosh~\cite{single-source} with the goal of inducing single-source robustness to \emph{random noise} in classification problems.
Our goal now is to study this hypothesis experimentally in 2D object detection domain.

\vspace{-0.4cm}
\begin{figure}[h]
    \centering
    \begin{subfigure}{0.235\textwidth}
    \centering
    \captionsetup{justification=centering}
    \includegraphics[width=\textwidth]{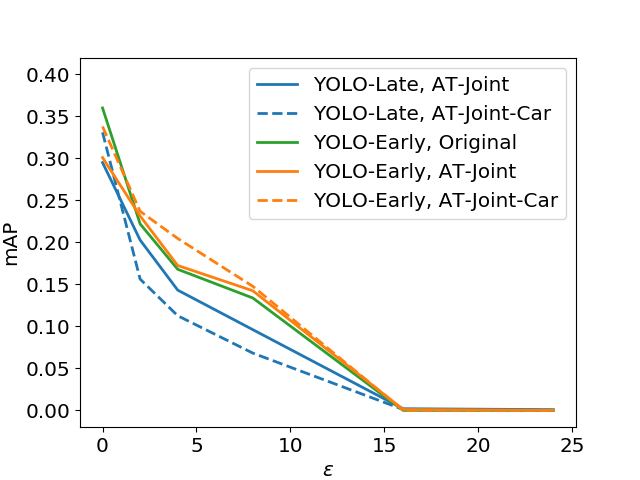}
    \caption{full-image attack}\label{fig:og-clean}
    \end{subfigure}
    \begin{subfigure}{0.235\textwidth}
    \centering
    \captionsetup{justification=centering}
    \includegraphics[width=\textwidth]{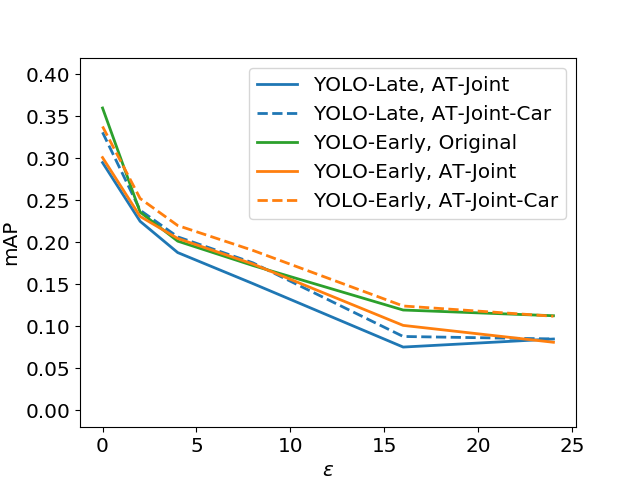}
    \caption{car-image attack}\label{fig:og-advi}
    \end{subfigure}
    \begin{subfigure}{0.235\textwidth}
    \centering
    \captionsetup{justification=centering}
    \includegraphics[width=\textwidth]{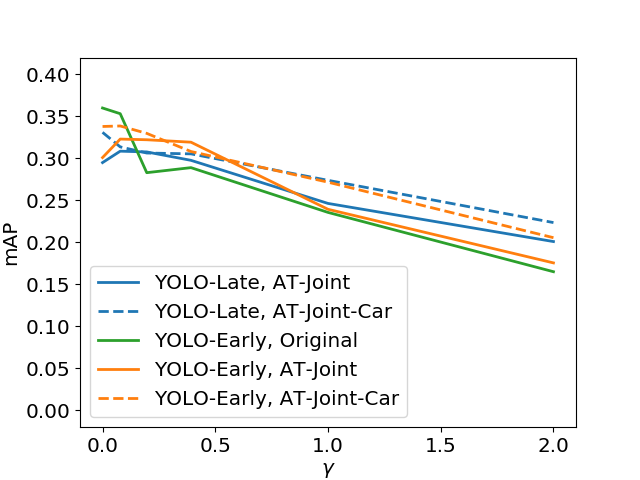}
    \caption{full-LiDAR attack}\label{fig:og-advc}
    \end{subfigure}
    \begin{subfigure}{0.235\textwidth}
    \centering
    \captionsetup{justification=centering}
    \includegraphics[width=\textwidth]{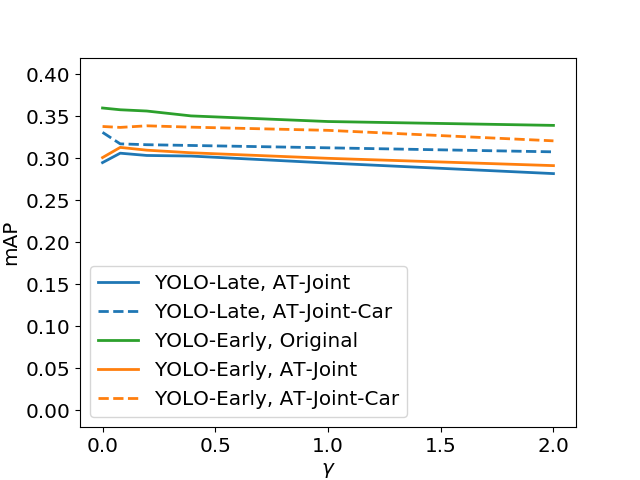}
    \caption{car-LiDAR attack}\label{fig:ati-clean}
    \end{subfigure}
    \caption{Early fusion v.s. late fusion with Joint-Channel Adversarial Training.}
    \label{fig:AT-joint}
\end{figure}

Figure~\ref{fig:AT-joint} presents the results on the efficacy of joint-channel adversarial training (Joint-AT).
Indeed, we see here a much improved picture compared to our results with single-channel adversarial training above.
For example, when facing full-image and full-LiDAR attacks, early fusion AT-Joint models now indeed outperform the pre-AT baseline in terms of robustness, although one has to acknowledge that this improvement isn't especially striking.
However, improvement is also not universal even here.
For example, late fusion still under-performs the original fusion model on image attacks (Figure~\ref{fig:AT-joint}(a) and (b)).
While all AT fusion models appear more robust in full-LiDAR attack evaluation (Figure~\ref{fig:AT-joint}(c)), they are all \emph{less} robust compared to original fusion model when the LiDAR attack is restricted to cars' bounding boxes (Figure~\ref{fig:AT-joint}(d)).
Finally, and somewhat surprisingly, now joint-channel training using attacks restricted only within cars' bounding boxes almost always outperform alternative Joint-AT approaches, \emph{even in robustness to full-image and full-LiDAR attacks}.
This last observation is very surprising, but overall serves to bolster our general observation that unrestricted conventional digital adversarial example models are not necessarily useful in developing robust classifiers.



\section{Conclusion}
\label{sec:conclusion}
We presented a systematic analysis of comparative robustness of deep fusion architectures involving RGB image and LiDAR channels.
Our first finding is that sensor fusion models are more robust against single-channel adversarial perturbation attacks than single-channel models before and after adversarial training, highlighting the role of fusion in improving robustness. 
Second, adversarial training on a single channel is problematic for sensor fusion models - it not only overfits with the background perturbation, but also makes the other sensor less robust. 
Third, jointly training on both sensors can mitigate the problem of single-sensor AT. 
Nevertheless, even joint-channel training fails to significantly improve robustness to single-channel attacks.
Fourth, we find throughout that the nature of the attack really matters when it comes to adversarial training.
Indeed, it is typically the case that adversarial training using perturbations to the entire input (say, every pixel in the input image) often overfits to the attack, and usually performs worse than fusion models before adversarial training.
Since such abstractions of attacks are not particularly well motivated in autonomous driving security in any case, our results suggest that threat modeling \emph{specifically for the purpose of defense} must be a critical area for future research in robust sensor fusion.


\section*{Acknowledgement}
This research was partially supported by the NSF (IIS-1905558, ECCS-2020289) and ARO (W911NF1910241).

{\small
\bibliographystyle{ieee_fullname}
\bibliography{egbib}
}

\begin{appendix}
\section{More Details of Training YOLOv4 Models on Waymo Dataset}
\paragraph{Normal and adversarial training}

YOLOv4 applies Bag of Freebies and Bag of Specials to improve performance. In this work, we only deploy Mosaic data augmentation for inputs. We use cross-stage partial connections (CSP) backbones and path augmentation network (PAN) necks with Mish activation. For all the clean YOLO models, we use SGD optimizer with learning rate from 0.01 to 0.002 with cosine annealing scheduler for 300 epochs. During training, we first pad the original image inputs to square shape, then resize them to $640\times 640$. Same procedure is done for dense depth map. We also normalize the depth maps over the entire training set. 

We use MMDetection~\cite{mmdet} as the basic detection framework for all YOLOv4 experiments. Details of normal training is already provided in the main text. For adversarial training, we use cyclic learning rate peaking at 40\% of the entire 25 epochs. The maximum learning rate is set to $10^{-3}$ for all AT variants. We also tried normal learning rate scheduling, i.e. decreasing from $10^{-3}$ until convergence, which takes much longer to finish, but we did not see any improvement in the robustness of models.
\paragraph{Depth map generation}
Here, we explain how we generate depth maps for RGB-D inputs. Waymo dataset has already provided camera projections of LiDAR point clouds, so we pre-generate and store the dense depth images and distance maps to speed up the normal training process. For adversarial attacks and adversarial training, the depth maps changes in every iteration/epoch of the process. Therefore, we use faiss~\cite{faiss}, a differentiable similarity search package, for nearest-neighbor interpolation. This is the fastest method we could find to generate accurate dense depth maps. Future directions include using depth estimation networks to trade accuracy for efficiency.

\begin{table*}[h]
\centering
\begin{tabular}{c|lllllll}
\hline
Models & Original & AT-Image & AT-Car & AT-LiDAR & AT-LiDAR-Car & AT-Joint & AT-Joint-Car \\ \hline
\hline
YOLO-RGB                     & 0.441    & 0.346    & 0.380        & -        & -            & -        & -            \\ \hline
YOLO-Depth                   & 0.323    & -        & -            & 0.259    & 0.274        & -        & -            \\ \hline
YOLO-Early                   & 0.478    & 0.411    & 0.441        & 0.398    & 0.412        & 0.424    & 0.446              \\ \hline
YOLO-Late                    & 0.450    & 0.403    & 0.422        & 0.403    & 0.417        & 0.405    &  0.428           \\ \hline
$AVOD^*$                       & 0.744    & 0.685    & 0.709        & 0.653    & 0.699        & 0.672 &  0.713           \\ \hline
\end{tabular}
\caption{Performance (mAP) of all models on clean (unperturbed) data before and after adversarial training.$^*$AVOD evaluates mAP on car class only.}
\label{T:cleanDataKitti}
\vspace{-0.1in}
\end{table*}

\begin{figure}[h]
    \centering
    \begin{subfigure}{0.235\textwidth}
    \centering
    \captionsetup{justification=centering}
    \includegraphics[width=\textwidth]{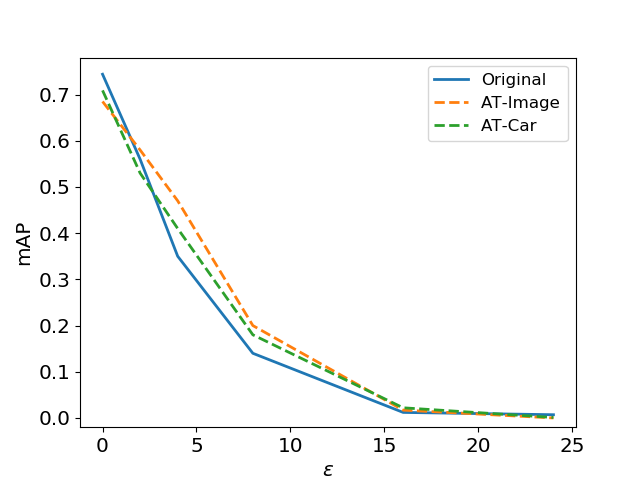}
    \caption{full-image attack}\label{fig:og-clean}
    \end{subfigure}
    \begin{subfigure}{0.235\textwidth}
    \centering
    \captionsetup{justification=centering}
    \includegraphics[width=\textwidth]{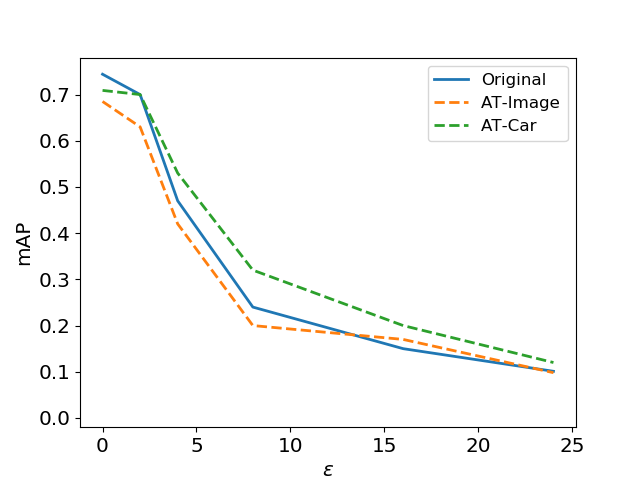}
    \caption{car-image attack}\label{fig:og-advi}
    \end{subfigure}
    \begin{subfigure}{0.235\textwidth}
    \centering
    \captionsetup{justification=centering}
    \includegraphics[width=\textwidth]{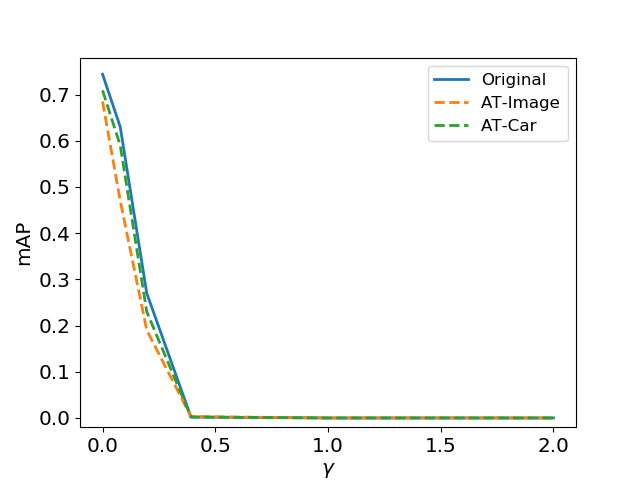}
    \caption{full-LiDAR attack}\label{fig:og-advc}
    \end{subfigure}
    \begin{subfigure}{0.235\textwidth}
    \centering
    \captionsetup{justification=centering}
    \includegraphics[width=\textwidth]{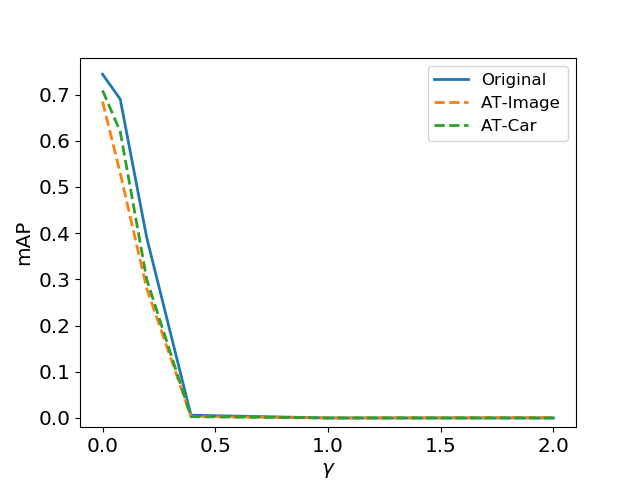}
    \caption{car-LiDAR attack}\label{fig:ati-clean}
    \end{subfigure}
    \caption{Comparing AT-Image variants for AVOD.}
    \label{fig:avod-AT-image}
\end{figure}

\begin{figure}[h]
    \centering
    \begin{subfigure}{0.235\textwidth}
    \centering
    \captionsetup{justification=centering}
    \includegraphics[width=\textwidth]{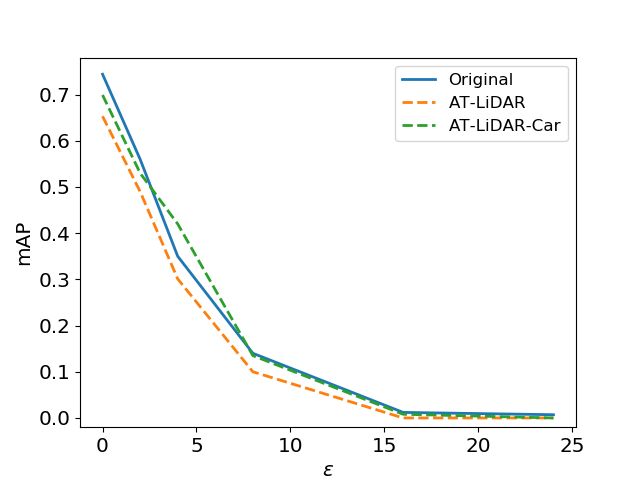}
    \caption{full-image attack}\label{fig:og-clean}
    \end{subfigure}
    \begin{subfigure}{0.235\textwidth}
    \centering
    \captionsetup{justification=centering}
    \includegraphics[width=\textwidth]{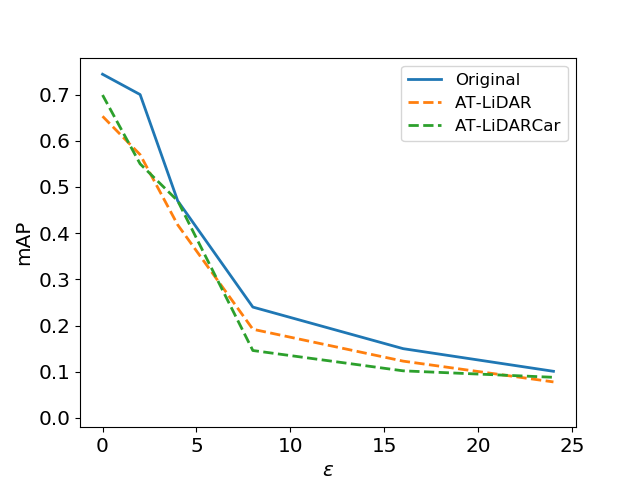}
    \caption{car-image attack}\label{fig:og-advi}
    \end{subfigure}
    \begin{subfigure}{0.235\textwidth}
    \centering
    \captionsetup{justification=centering}
    \includegraphics[width=\textwidth]{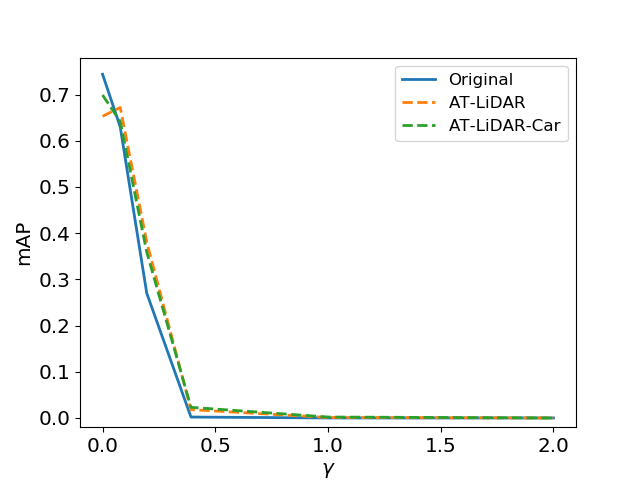}
    \caption{full-LiDAR attack}\label{fig:og-advc}
    \end{subfigure}
    \begin{subfigure}{0.235\textwidth}
    \centering
    \captionsetup{justification=centering}
    \includegraphics[width=\textwidth]{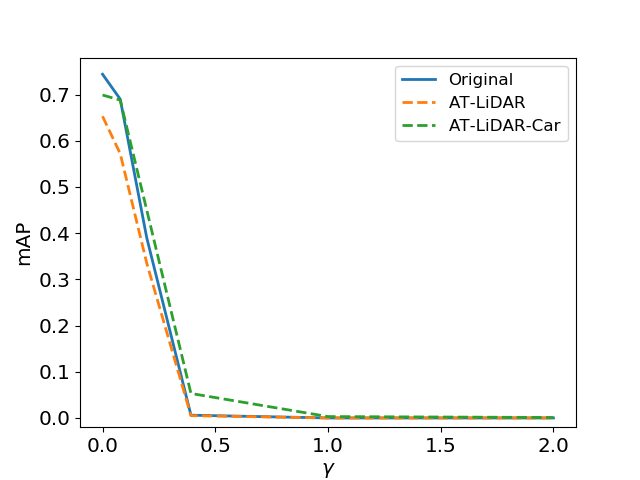}
    \caption{car-LiDAR attack}\label{fig:ati-clean}
    \end{subfigure}
    \caption{Comparing AT-LiDAR variants for AVOD.}
    \label{fig:avod-AT-lidar}
\end{figure}

\begin{figure}[h]
    \centering
    \begin{subfigure}{0.235\textwidth}
    \centering
    \captionsetup{justification=centering}
    \includegraphics[width=\textwidth]{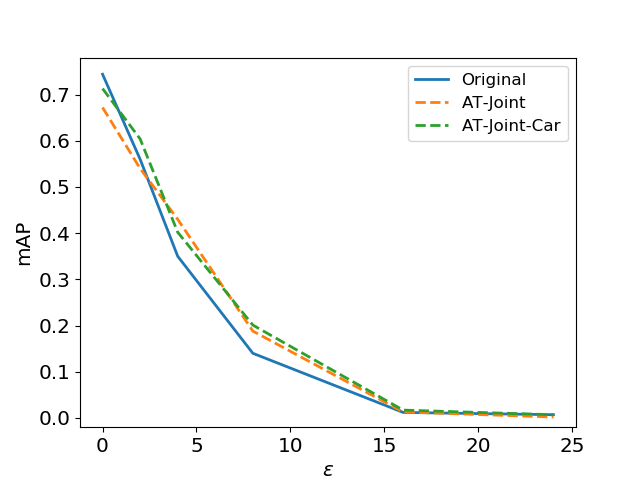}
    \caption{full-image attack}\label{fig:og-clean}
    \end{subfigure}
    \begin{subfigure}{0.235\textwidth}
    \centering
    \captionsetup{justification=centering}
    \includegraphics[width=\textwidth]{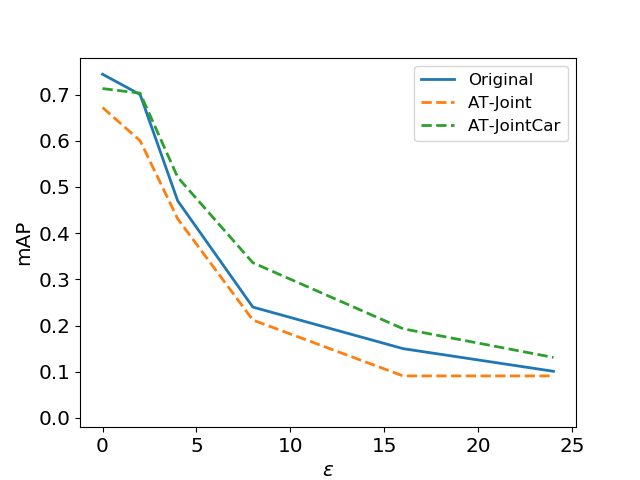}
    \caption{car-image attack}\label{fig:og-advi}
    \end{subfigure}
    \begin{subfigure}{0.235\textwidth}
    \centering
    \captionsetup{justification=centering}
    \includegraphics[width=\textwidth]{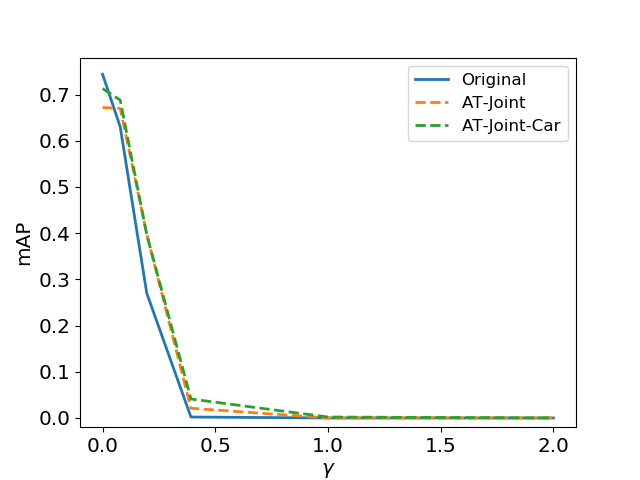}
    \caption{full-LiDAR attack}\label{fig:og-advc}
    \end{subfigure}
    \begin{subfigure}{0.235\textwidth}
    \centering
    \captionsetup{justification=centering}
    \includegraphics[width=\textwidth]{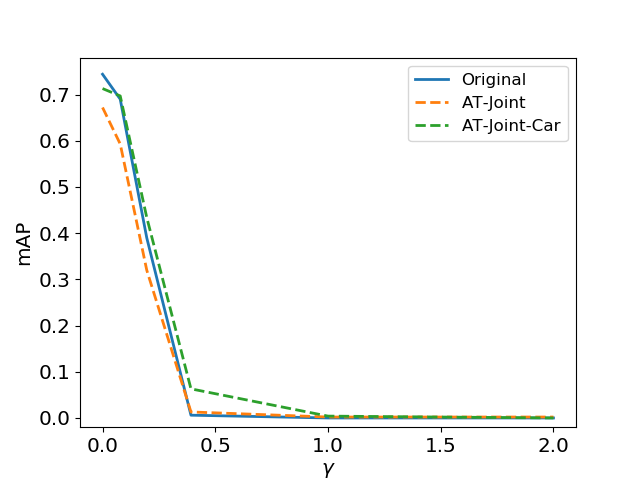}
    \caption{car-LiDAR attack}\label{fig:ati-clean}
    \end{subfigure}
    \caption{Comparing AT-Joint variants for AVOD.}
    \label{fig:avod-AT-joint}
\end{figure}

\section{Experiments on KITTI Dataset}
\subsection{Experiment Settings}
We migrate all the experiments for YOLOv4 onto the KITTI dataset.
We also run part of the experiments for AVOD on KITTI. As mentioned in the main text, AVOD is a two-stage camera-LiDAR fusion detector. Identifying single-sensor equivalent of AVOD is not as straightforward as we did for YOLO, so we only try attacks and AT on the fusion model itself.
We train YOLO models using exactly the same settings as we used on Waymo. For AVOD, we use the settings in their paper for normal training, and a learning rate of $5\times 10^{-5}$ for AT with normal learning rate scheduling. We did not use cyclic learning rate here because the peak learning rate is too small and AT on AVOD is much faster than on YOLO. We use $\epsilon=2$ for AT-Image variants of AVOD, but $\gamma=0.15$, smaller than that for YOLO, for AT-LiDAR and AT-Joint variants because AVOD is much more vulnerable to LiDAR attacks than YOLO models using depth maps. We will discuss this up next.

For KITTI dataset, we use the default train-validation split provided by the official site which gives us 3,721 samples, and randomly sample 118 inputs from the validation set to test our attacks.

\subsection{Results}
The AVOD model only detects car class, while YOLOv4 models predict all three classes, and we train each YOLO model using its corresponding pretrained weights on Waymo dataset. The mAP scores of all models (including the AT variants) on clean data are shown in Table~\ref{T:cleanDataKitti}.

Figures~\ref{fig:avod-AT-image},~\ref{fig:avod-AT-lidar},~\ref{fig:avod-AT-joint} illustrate the experiment results for AVOD. We can clearly see that both channels are very vulnerable to adversarial attacks, and LiDAR channel is even more vulnerable. 
AVOD uses bird's eye view (BEV) map instead of depth map for LiDAR representation, which could be the reason why LiDAR channel of AVOD is vulnerable. Per-point perturbation on LiDAR is completely preserved through the BEV generation process, but is partially smoothed out during the generation of depth maps because of the projection and interpolation steps. 
This is the reason why we chose a smaller $\gamma$ for AT-LiDAR and AT-Joint on AVOD. After AT on each channel respectively, we can see a slight increase in the robustness of the target channel. Still, the robustness of the other channel is decreased. After AT on both channels jointly, we notice from Figure~\ref{fig:avod-AT-joint} that although the robustness of LiDAR channel does not change much from AT-LiDAR, the robustness of image channel increased dramatically comparing to the results in Figure~\ref{fig:avod-AT-lidar}.

Figures~\ref{fig:imgc} - \ref{fig:AT-joint} show the results for YOLO variants on KITTI. The conclusions are mostly the same as what we obtained on Waymo dataset, which means that the phenomenon we observed is not an issue of dataset, but a general problem for the YOLO sensor fusion models. One noticeable difference is that YOLO-Early is extremely robust against LiDAR attacks, which seems that it is making even less use of LiDAR when compared to YOLO-Early on Waymo.

Note that Kim and Ghosh~\cite{single-source} suggest that to achieve single-source robustness for fusion models, we should AT on both sensors jointly, but our results suggest that single-source robustness can be achieved by AT on that sensor or jointly on both sensors, but to guarantee the overall robustness of fusion models, we do need joint AT on both sensors.

\begin{figure}[h]
    \centering
    \begin{subfigure}{0.235\textwidth}
    \centering
    \captionsetup{justification=centering}
    \includegraphics[width=\textwidth]{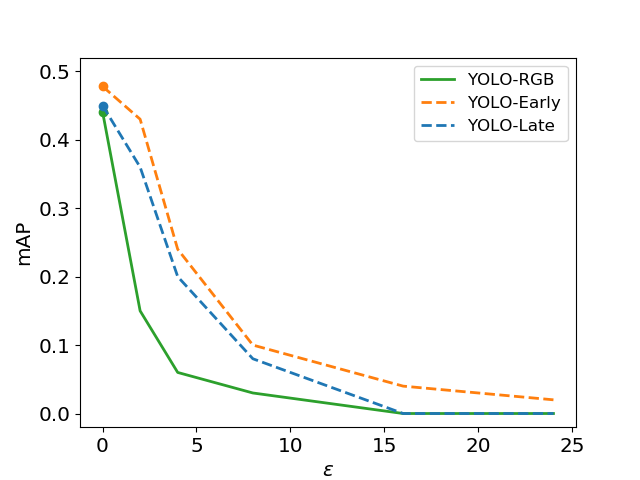}
    \caption{full-image attack}\label{fig:yolo-og-imageatk}
    \end{subfigure}
    \begin{subfigure}{0.235\textwidth}
    \centering
    \captionsetup{justification=centering}
    \includegraphics[width=\textwidth]{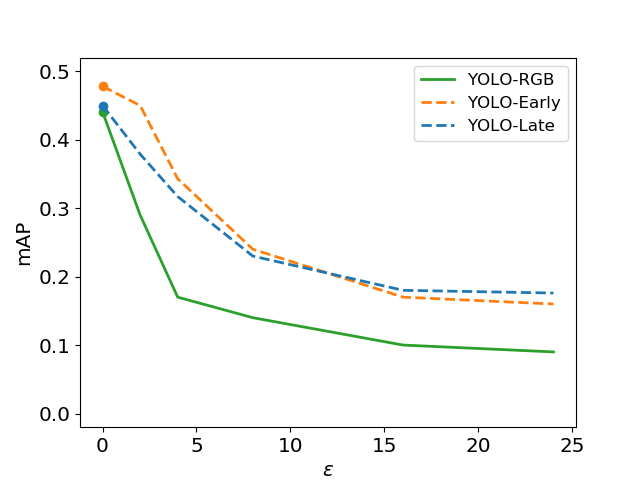}
    \caption{car-image attack}\label{fig:yolo-og-carimg}
    \end{subfigure}
    \begin{subfigure}{0.235\textwidth}
    \centering
    \captionsetup{justification=centering}
    \includegraphics[width=\textwidth]{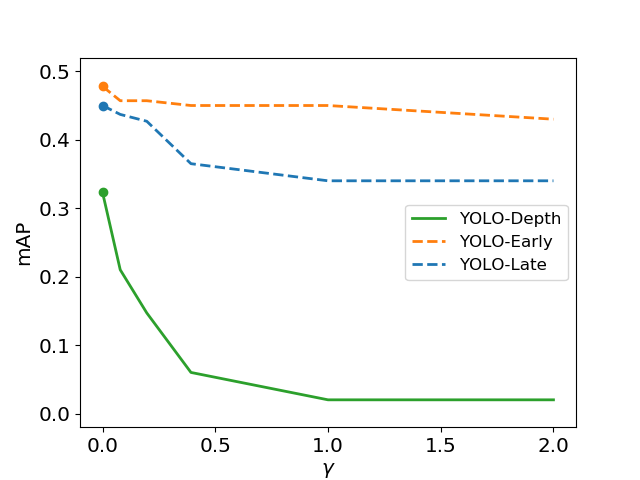}
    \caption{full-LiDAR attack}\label{fig:yolo-og-lidaratk}
    \end{subfigure}    
    \begin{subfigure}{0.235\textwidth}
    \centering
    \captionsetup{justification=centering}
    \includegraphics[width=\textwidth]{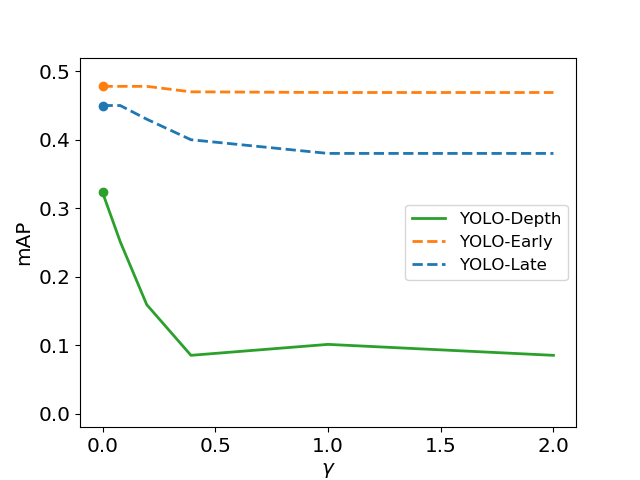}
    \caption{car-LiDAR attack}\label{fig:yolo-og-carlidaratk}
    \end{subfigure}
    \caption{Robustness of deep sensor fusion, compared to single-channel neural network object detection models, to adversarial attacks on image and LiDAR modalities.} 
    \label{fig:imgc}
\end{figure}

\begin{figure}[h]
\centering
    \begin{subfigure}{0.235\textwidth}
    \centering
    \captionsetup{justification=centering}
    \includegraphics[width=\textwidth]{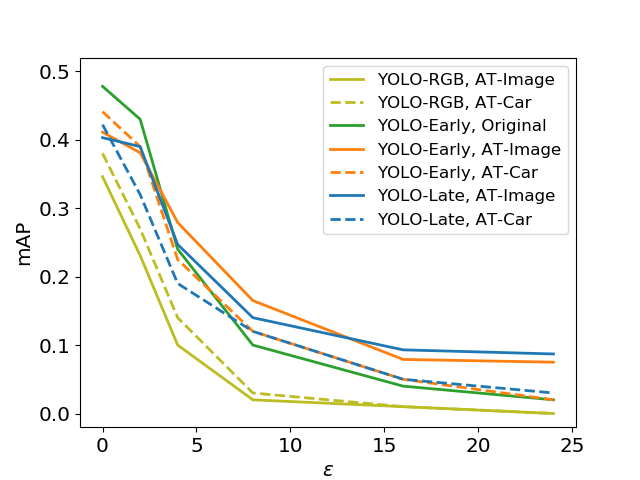}
    \end{subfigure}
    \begin{subfigure}{0.235\textwidth}
    \centering
    \captionsetup{justification=centering}
    \includegraphics[width=\textwidth]{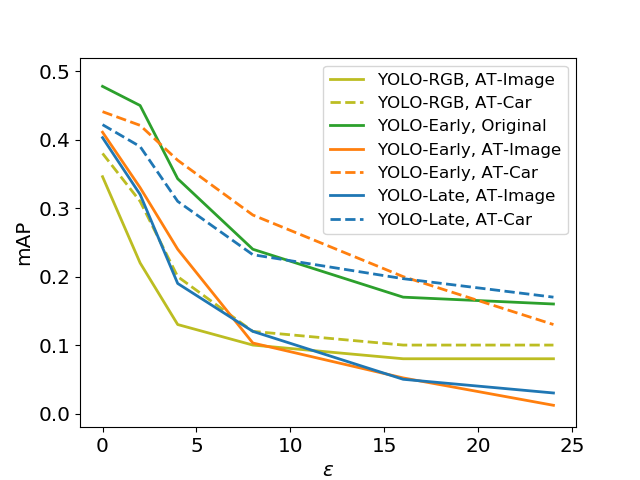}
    \end{subfigure}
    \caption{Early and late fusion after adversarial training, and under attacks, involving only the image channel.  Left: full-image attack.  Right: car bounding box attack.}
    \label{fig:AT-image}
\end{figure}

\begin{figure}[h]
\centering
    \begin{subfigure}{0.235\textwidth}
    \centering
    \captionsetup{justification=centering}
    \includegraphics[width=\textwidth]{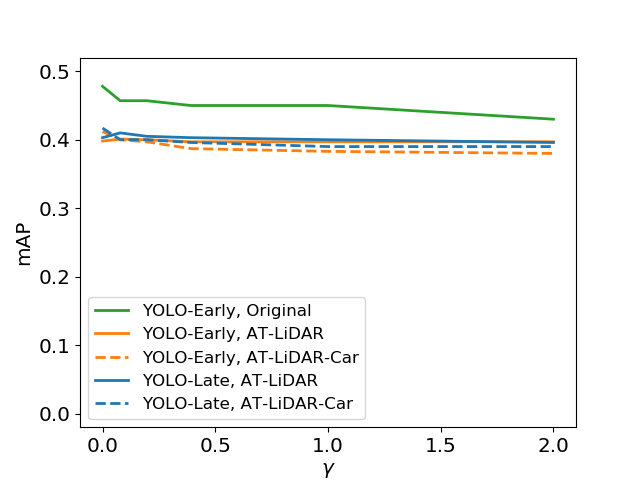}
    \end{subfigure}
    \begin{subfigure}{0.235\textwidth}
    \centering
    \captionsetup{justification=centering}
    \includegraphics[width=\textwidth]{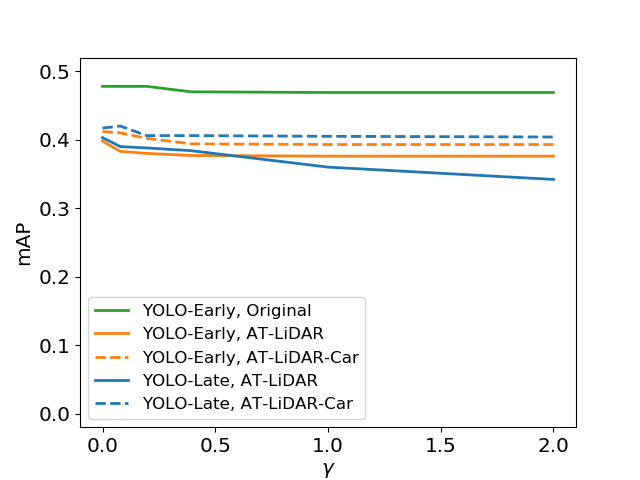}
    \end{subfigure}
    \caption{AT with LiDAR attacks, and LiDAR Channel Attacks. Left: full-LiDAR attack. Right: car-LiDAR attack.}
    \label{fig:AT-lidar}
\end{figure}

\begin{figure}[h]
\centering
    \begin{subfigure}{0.235\textwidth}
    \centering
    \captionsetup{justification=centering}
    \includegraphics[width=\textwidth]{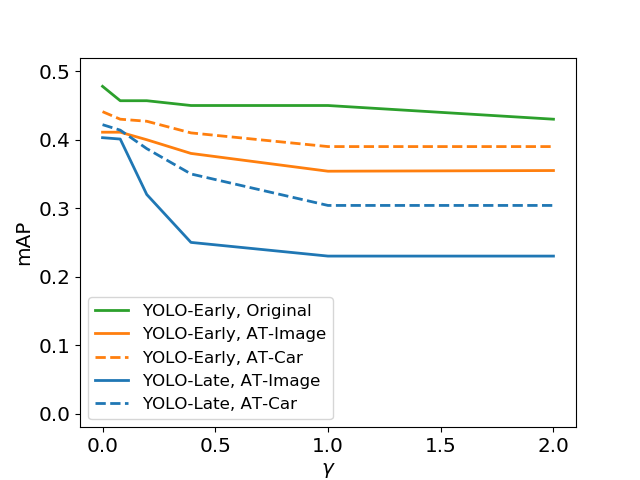}
    \label{fig:at-fulllidar}
    \end{subfigure}
    \begin{subfigure}{0.235\textwidth}
    \centering
    \captionsetup{justification=centering}
    \includegraphics[width=\textwidth]{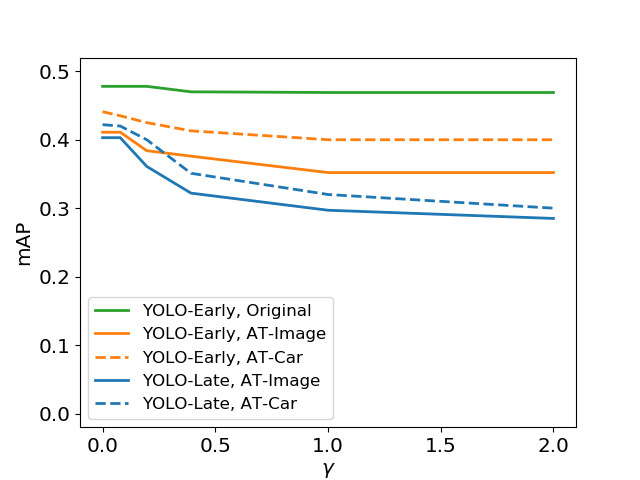}
    \label{fig:at-carlidar}
    \end{subfigure}
    \caption{Early and late fusion after adversarial training using image attacks, but subject to LiDAR attacks. Left: full-LiDAR attack. Right: car-LiDAR attack.}
\label{fig:ATI-lidar}
\end{figure}

\begin{figure}[h]
\centering
    \begin{subfigure}{0.235\textwidth}
    \centering
    \captionsetup{justification=centering}
    \includegraphics[width=\textwidth]{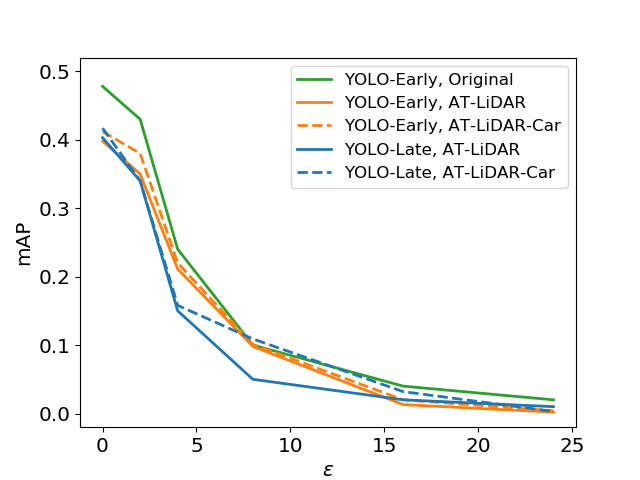}
    \end{subfigure}
    \centering
    \begin{subfigure}{0.235\textwidth}
    \centering
    \captionsetup{justification=centering}
    \includegraphics[width=\textwidth]{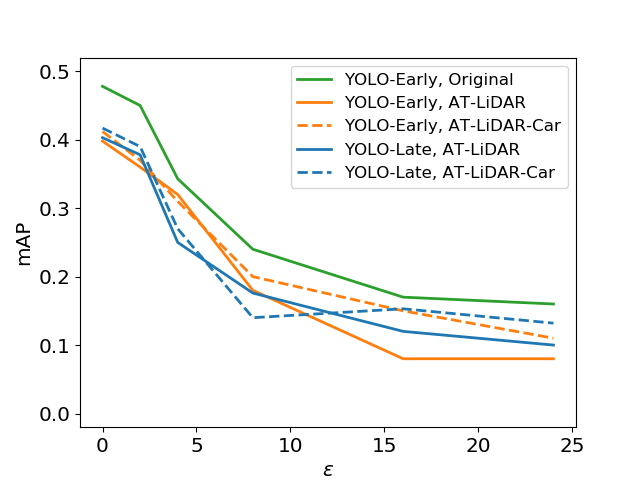}
    \end{subfigure}
    \caption{Early and late fusion after adversarial training using LiDAR attacks, but subject to image attacks. Left: full-image attack. Right: car-image attack.}
\label{fig:ATL-image}
\end{figure}

\begin{figure}[h]
    \centering
    \begin{subfigure}{0.235\textwidth}
    \centering
    \captionsetup{justification=centering}
    \includegraphics[width=\textwidth]{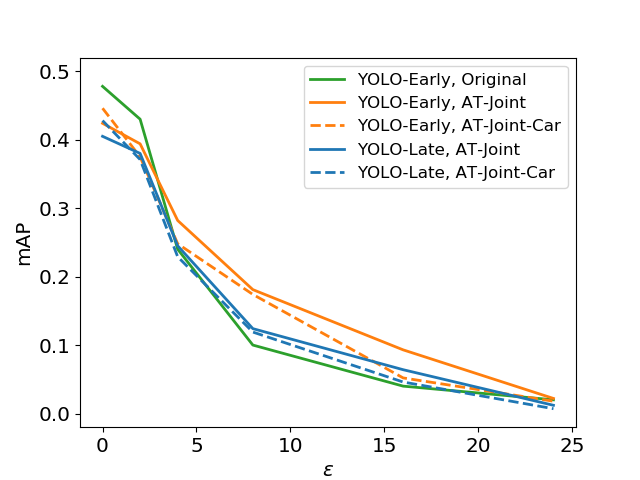}
    \caption{full-image attack}\label{fig:og-clean}
    \end{subfigure}
    \begin{subfigure}{0.235\textwidth}
    \centering
    \captionsetup{justification=centering}
    \includegraphics[width=\textwidth]{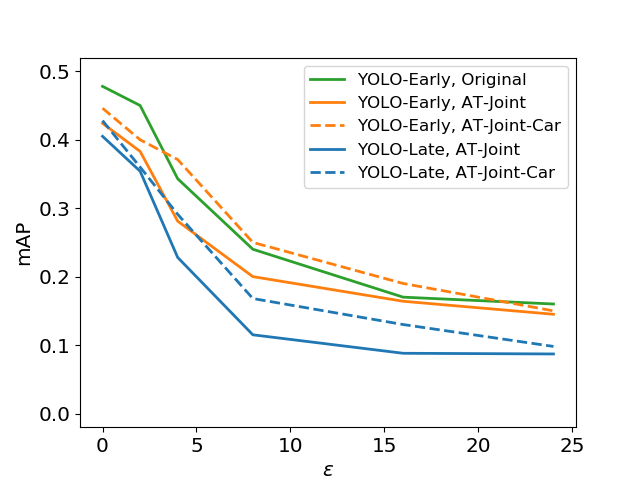}
    \caption{car-image attack}\label{fig:og-advi}
    \end{subfigure}
    \begin{subfigure}{0.235\textwidth}
    \centering
    \captionsetup{justification=centering}
    \includegraphics[width=\textwidth]{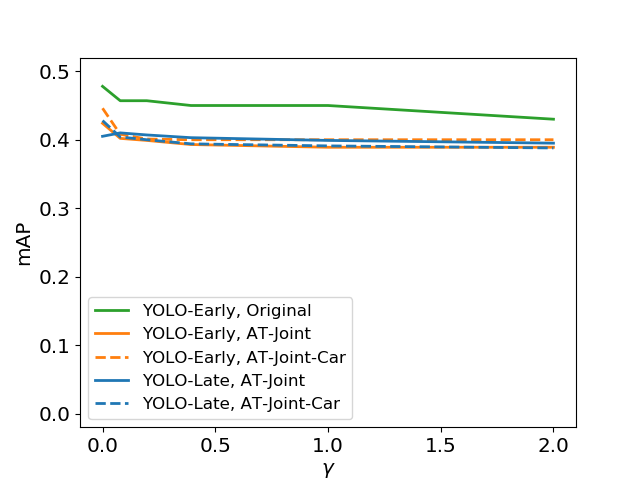}
    \caption{full-LiDAR attack}\label{fig:og-advc}
    \end{subfigure}
    \begin{subfigure}{0.235\textwidth}
    \centering
    \captionsetup{justification=centering}
    \includegraphics[width=\textwidth]{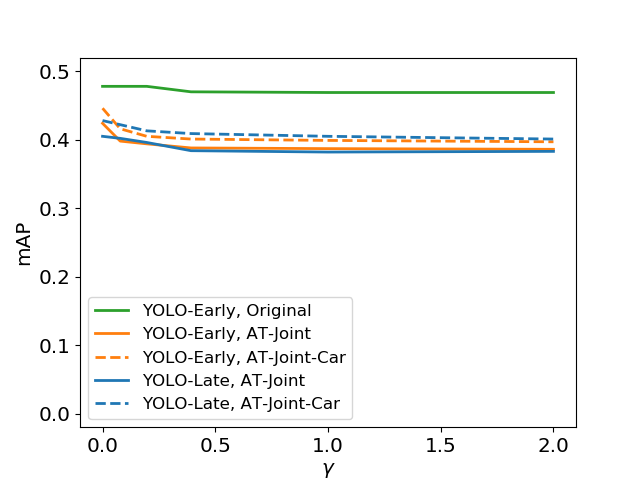}
    \caption{car-LiDAR attack}\label{fig:ati-clean}
    \end{subfigure}
    \caption{Early fusion v.s. late fusion with Joint-Channel Adversarial Training.}
    \label{fig:AT-joint}
\end{figure}

\section{Black-box Attacks}
In addition to white-box attacks, we perform black-box attacks on sensor fusion models. In detail, we generate white-box perturbations on model A, and apply the perturbations directly on model B for evaluation. All the experiment results in this section are obtained on Waymo dataset.

\begin{figure}[h]
    \centering
    \begin{subfigure}{0.235\textwidth}
    \centering
    \captionsetup{justification=centering}
    \includegraphics[width=\textwidth]{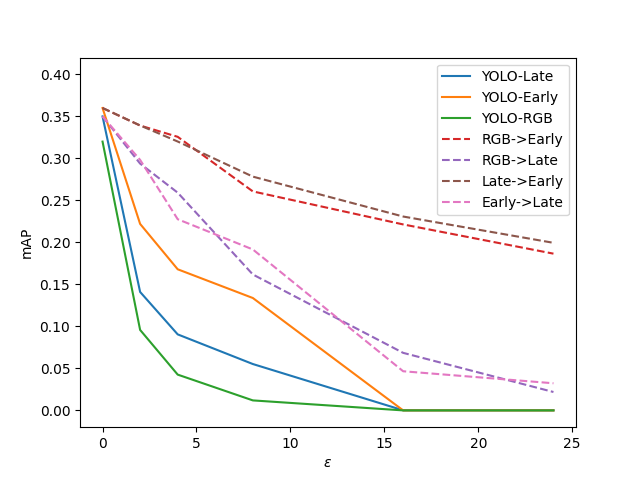}
    \caption{full-image attack}\label{fig:og-clean}
    \end{subfigure}
    \begin{subfigure}{0.235\textwidth}
    \centering
    \captionsetup{justification=centering}
    \includegraphics[width=\textwidth]{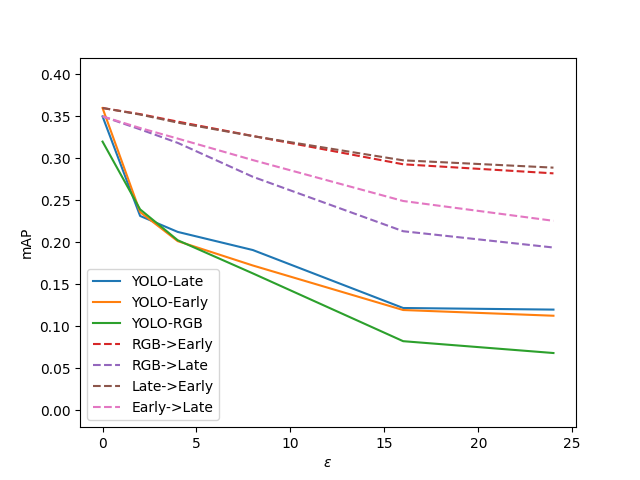}
    \caption{car-image attack}\label{fig:og-advi}
    \end{subfigure}
    \begin{subfigure}{0.235\textwidth}
    \centering
    \captionsetup{justification=centering}
    \includegraphics[width=\textwidth]{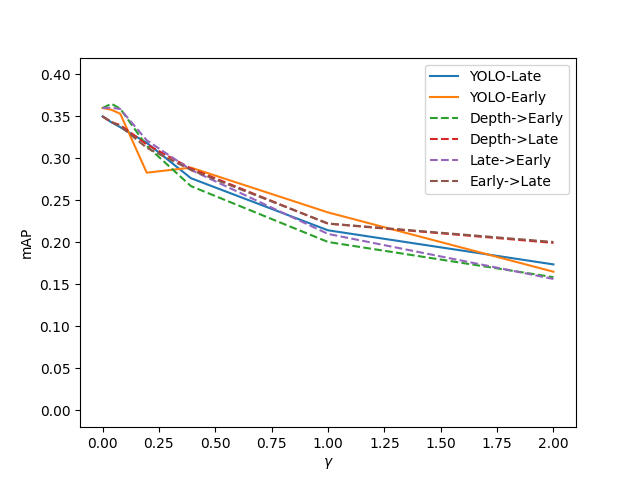}
    \caption{full-LiDAR attack}\label{fig:og-advc}
    \end{subfigure}
    \begin{subfigure}{0.235\textwidth}
    \centering
    \captionsetup{justification=centering}
    \includegraphics[width=\textwidth]{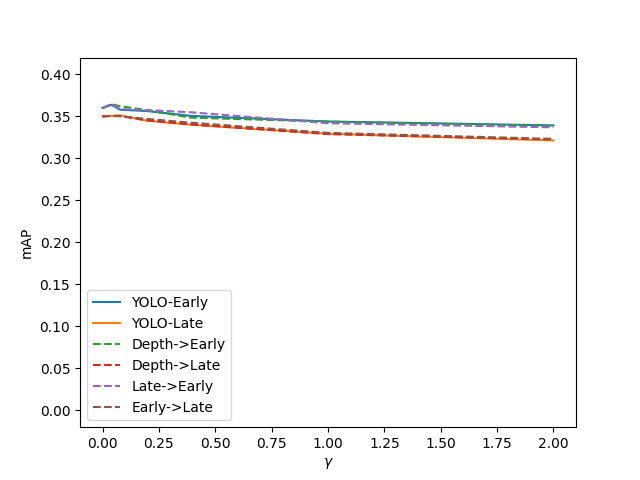}
    \caption{car-LiDAR attack}\label{fig:ati-clean}
    \end{subfigure}
    \caption{Black-box attacks across different models. Black-box attack results are in dashed lines. The arrows in legend denote the perturbation generated on the left-hand-side model and applied on the right-hand-side model.} 
    \label{fig:bbox}
\end{figure}

Figure~\ref{fig:bbox} shows how black-box attacks work on fusion models. While we can notice black-box attacks are inferior to white-box attacks for image channel, black-box attacks on LiDAR channel is as ineffective as white-box attacks.

\begin{figure}[h]
    \centering
    \begin{subfigure}{0.235\textwidth}
    \centering
    \captionsetup{justification=centering}
    \includegraphics[width=\textwidth]{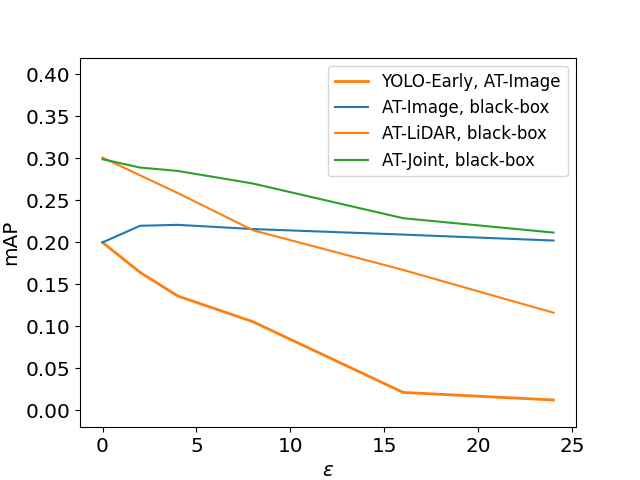}
    \caption{full-image attack}\label{fig:og-clean}
    \end{subfigure}
    \begin{subfigure}{0.235\textwidth}
    \centering
    \captionsetup{justification=centering}
    \includegraphics[width=\textwidth]{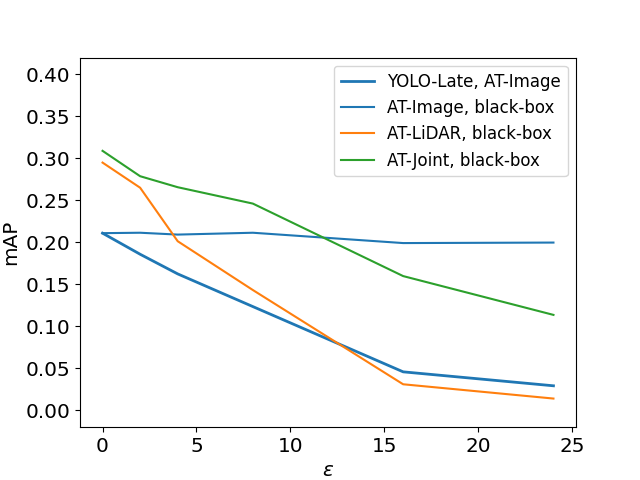}
    \caption{car-image attack}\label{fig:og-advi}
    \end{subfigure}
    \begin{subfigure}{0.235\textwidth}
    \centering
    \captionsetup{justification=centering}
    \includegraphics[width=\textwidth]{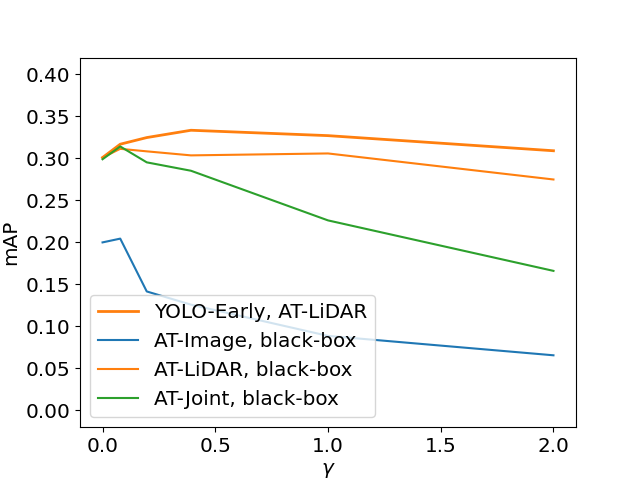}
    \caption{full-LiDAR attack}\label{fig:og-advc}
    \end{subfigure}
    \begin{subfigure}{0.235\textwidth}
    \centering
    \captionsetup{justification=centering}
    \includegraphics[width=\textwidth]{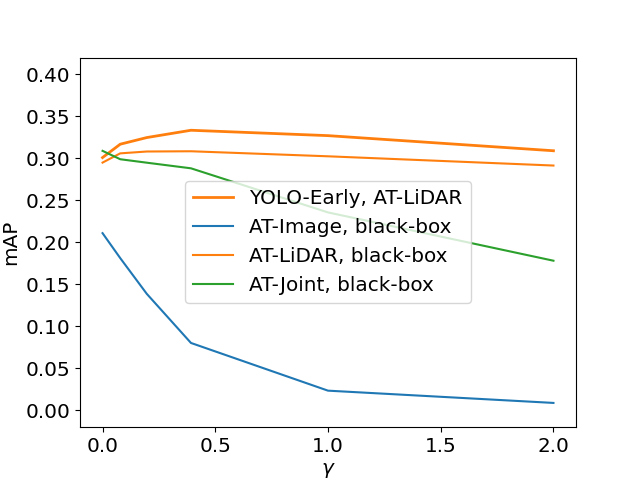}
    \caption{car-LiDAR attack}\label{fig:ati-clean}
    \end{subfigure}
    \caption{Black-box attacks after white-box AT. The arrows in legend denote the perturbation generated on the left-hand-side model and applied on the right-hand-side model.} 
    \label{fig:bbox-after-at}
\end{figure}

Figure~\ref{fig:bbox-after-at} shows how white-box AT affects the effectiveness of black-box attacks. All perturbations are generated on the corresponding single-sensor model, e.g. in full-image black-box attacks, we generate perturbation on YOLO-RGB after AT-Image, and apply the perturbation on YOLO-Early model after AT-Image. In (a) and (b), AT-Image brings clear improvement on model robustness against black-box image attacks, though we cannot observe such effect for AT-LiDAR.

We also attempt to adversarially train the model with black-box perturbations, i.e. black-box adversarial training. Specifically, we generate perturbation on YOLO-RGB model, and use the perturbation for adversarial training on the image channel of YOLO-Early model, and then evaluate black-box image attacks on YOLO-Early with perturbation generated from YOLO-Late model. However, the model performance of YOLO-Early on clean data is largely sabotaged after such training process. This can be left as future work where researchers can try black-box AT on simpler tasks like classification.

\end{appendix}

\end{document}